\documentclass{article}
\usepackage{iclr2027_conference,times}

\newif\ifshowrefute

\showrefutefalse

\usepackage{amsmath,amsfonts,bm}

\def\eqref#1{equation~\ref{#1}}

\def\1{\bm{1}}

\DeclareMathAlphabet{\mathsfit}{\encodingdefault}{\sfdefault}{m}{sl}
\SetMathAlphabet{\mathsfit}{bold}{\encodingdefault}{\sfdefault}{bx}{n}

\usepackage{amsmath}
\usepackage{array}
\usepackage{booktabs}
\usepackage{wrapfig}
\usepackage{graphicx}
\usepackage{hyperref}
\usepackage{microtype}
\usepackage{url}
\usepackage{multirow}
\usepackage{placeins}
\usepackage{tikz}
\graphicspath{{figures/}}
\usetikzlibrary{positioning,arrows.meta,calc,backgrounds}

\title{The Key Handoff:\\Retrieval in Hybrid Language Models}

\author{Kaan Kale$^{1}$, Oguzhan Baser$^{2}$, Sriram Vishwanath$^{1}$ \\
\normalfont $^{1}$Georgia Institute of Technology, USA \\
\normalfont $^{2}$University of Texas at Austin, USA\\
\normalfont \texttt{hkale7@gatech.edu,  oguzhanbaser@utexas.edu, sriram@ece.gatech.edu}}

\iclrfinalcopy


\begin{document}
\maketitle
\lhead{}

\begin{abstract}
A two-hop question makes a language model retrieve twice: once to produce a
bridge entity, and once to retrieve with it.  Transformers resolve that
entity in their early layers.  Hybrid models replace most of the attention
with a recurrent state, so where the key becomes usable, and where it is
spent on an answer, is not known.  Answering ``Where is the ball?'' from
``the ball belongs to Alice'' and ``Alice is in the garden'' turns on
Alice, a name the question does not mention.  A model could reach garden
through Alice, the key it computed, or through where the fact sits in the
prompt. Across twelve models, dense and hybrid, we move a hidden state from one
story into another where the two routes lead to different places, and read
off which place the model gives. An attention layer converts the key in every model we
tested, and crossing that layer removes its usable effect, all of it where
no attention follows.  In sequential hybrids this makes the answer a
handoff: recurrent layers carry the key forward, and attention spends it.
Retrieval does not always stop there.  Writing a different fact into the
memory of the recurrent layers that follow the last attention layer can
move the answer toward that fact, multiplying its odds by \textbf{1.3} to
\textbf{2.7}, and which hybrids do this is not settled by their
architecture or training.  That read is addressed by the key, not by
position: recurrent state in a hybrid is not only a carrier, but a memory
that later layers can query.
\end{abstract}

\section{Introduction}
A ball belongs to Alice, and Alice is in the garden.  To answer
\emph{Where is the ball?}, a language model first resolves Alice, then uses
Alice as a \textit{retrieval key} to find the garden.  We ask where a
usable key emerges, where it becomes an answer, and which components
perform those operations.

Hybrid language models interleave attention, which accesses earlier token
representations, with recurrent computation, which maintains a state as it
reads the sequence
\citep{gu2024mamba,dao2024transformersssmsgeneralizedmodels,waleffe2024empiricalstudymambabasedlanguage,nvidia2025nemotronhfamilyaccurateefficient},
whereas dense transformers apply attention at every layer, so the same
two-hop task can be solved with the intermediate computation distributed
differently. For transformers this computation has a known shape: the intermediate
entity is resolved in the early layers
\citep{biran-etal-2024-hopping,yang-etal-2024-large-language-models}, and a
small set of attention heads does the retrieving \citep{wu2025retrieval}.
Hybrids remove most of that attention.  That is the point of the design: a recurrent layer holds one state instead
of every earlier token.  But attention is what prior work ties retrieval
to, and a hybrid has the least of it.
Work on them so far finds retrieval still leaning on the attention that
remains \citep{michalak2025some,bick2025understanding}. What the recurrent layers
contribute, and whether any of the work happens where no attention is left
to do it, is open.

\begin{figure}[t]
  \centering
  \setlength{\abovecaptionskip}{4pt}
  \begin{minipage}[b]{0.66\linewidth}
    \centering
    \includegraphics[width=\linewidth]{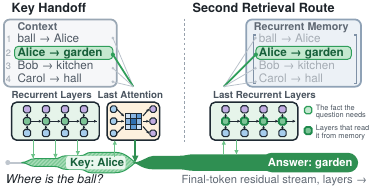}\\[2pt]
    {\small (a)}
  \end{minipage}\hfill
  \begin{minipage}[b]{0.28\linewidth}
    \centering
    \includegraphics[width=\linewidth,trim=50pt 70pt 49pt 60pt,clip]{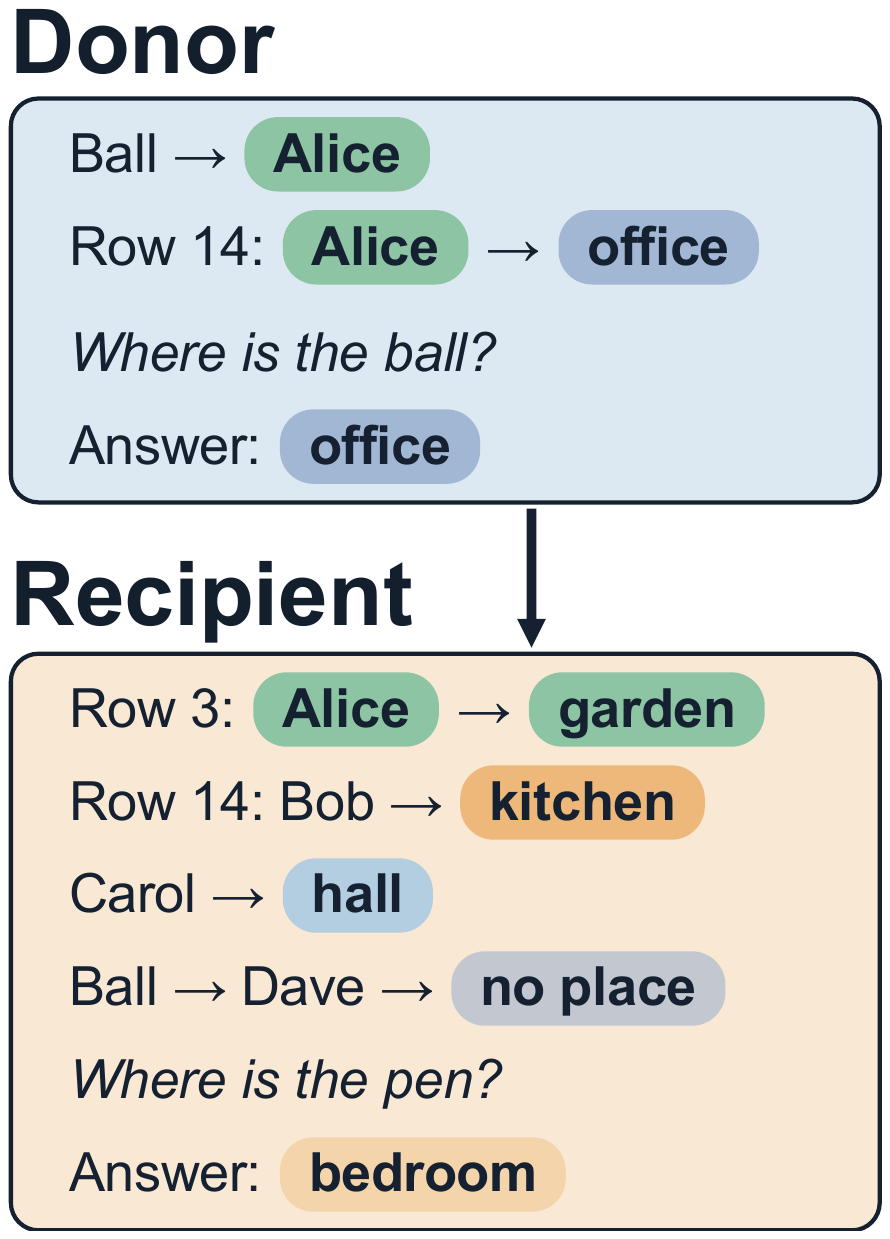}\\[2pt]
    {\small (b)}
  \end{minipage}\hfill
  \caption{\textbf{The handoff, and a pair that measures it.}  (a) In a
  sequential hybrid the final token's stream is hatched while it carries the
  key to an attention layer that reads it, and solid once it carries the
  answer; after the last attention layer a recurrent layer can retrieve a
  binding from its own memory.  (b) The two prompts share a row, so the
  donor's carried answer, that row, and the key it computes point at three
  different places; hall is predicted by none and serves as the reference.
  Only the donor's final-token state moves, into the same layer input of the
  recipient.  Names and places stand in for the arrow task's keys and
  numbers; the literal prompts are in Appendix~\ref{app:inventory}.}
  \label{fig:handoff}
  \label{fig:placeholder}
\end{figure}

Our \textbf{crossed assay} distinguishes a computed key from a carried
answer or a pointer to a fact's position by moving a state from one story
(the \emph{donor}) into another (the \emph{recipient}) whose facts are
arranged so that the three predict three different answers
(Figure~\ref{fig:handoff}b).  Building on counterfactual binding studies
\citep{feng2024how,gur-arieh2026mixing}, it varies the key and the
position of the donor's fact independently, and scores both against an
answer no account predicts. The key turns out to move through a sequential hybrid as a handoff:
recurrent layers carry it, an attention layer spends it, and retrieval does
not always stop there (Figure~\ref{fig:handoff}a). We reach that picture
with one measurement and three results:
\begin{itemize}\setlength{\itemsep}{0.1pt}
 \item \textbf{A measurement for computed keys.}  An activation-patching
  assay that follows a computed key across architectures and kinds of
  intervention, over 23 depth profiles from twelve checkpoints in nine
  model families (Section~\ref{sec:setup}).
 \item \textbf{Where keys become answers.}  In every profile the key peaks
  at the input of the layer that converts it, and crossing that layer
  removes the key's usable effect: all of it where that layer is the
  model's last attention layer, part of it where attention follows
  (Section~\ref{sec:profile}).
 \item \textbf{A division of labor in hybrids.}  In four hybrids recurrent
  layers carry the key to an attention layer, where a few heads resolve it
  against the facts in context (Section~\ref{sec:mechanism}).
 \item \textbf{Retrieval after attention.}  Writing a different fact into
  the recurrent memory that follows the last attention layer moves the
  answer toward it, and that read is addressed by the key rather than by
  position (Section~\ref{sec:suffix}).
\end{itemize}

\section{A crossed assay for computed-key retrieval}
\label{sec:setup}
\label{sec:constructions}

\paragraph{A donor and a recipient.}
In Figure~\ref{fig:handoff}b the \emph{donor} asks where the ball is: the
ball belongs to Alice, and the fact placing her in the office sits at row
14.  The \emph{recipient} asks where the pen is and answers bedroom; among
its 17 facts it places Alice in the garden, Bob in the kitchen at row 14,
and Carol in the hall.  The ball's owner has no stated location, so a
transplanted query token alone has nothing to resolve here
(Appendix~\ref{app:crossed}).  After the transplant a carried answer
predicts office, a computed Alice key predicts garden (\emph{key-based}
retrieval), and a row-14 pointer predicts kitchen (\emph{row-based}
retrieval, positional retrieval in the sense of
\citet{gur-arieh2026mixing}).  Carol's hall is predicted by none of the
three and serves as the \emph{reference candidate}.

\paragraph{The transplant.}
For donor \(x_d\) and recipient \(x_r\), let \(h_l\) be the \emph{residual
state} entering layer \(l\), the representation passed between layers.  At
the final prompt token we copy \(h_l(x_d)\) into the recipient at the same
layer input, and the recipient finishes its forward pass with its own
tokens and its computation over earlier tokens unchanged.  No learned
transformation is involved \citep{geiger2021causal}.  For each candidate
answer \(c\) we measure the change in its logit \(\ell\) against the same
recipient without intervention:
\begin{equation}
 h_l(x_r)\leftarrow h_l(x_d),\qquad
 \Delta\ell(c)=\ell\bigl(c\mid x_r,h_l(x_d)\bigr)-\ell(c\mid x_r).
 \label{eq:transplant}
\end{equation}
A positive change promotes that candidate.
A recurrent layer adds an output \(o_l\) to the residual \(h_l\), computing
it from \(h_l\) and a \emph{memory} \(m_l\) of the preceding tokens.  Every
intervention here replaces one of these at the final token: \(h_l\) in this
section, \(o_l\) and attention-head outputs in
Section~\ref{sec:mechanism}, and \(m_l\) in Section~\ref{sec:suffix}.

\paragraph{Separating the key from its row.}
Across pairs the donor's key and the row of its completing fact vary
independently, and garden, kitchen, and hall each take their turn in the
key, position, and reference roles, so a place the model simply prefers
cannot pass for key-based retrieval (Appendix
Table~\ref{tab:factorial}).  A generated story, with its own people, places
and fact order, together with the donor and recipient prompts built from
it, is one \emph{identity}; the experiments score 200 identities each
except where Appendix Table~\ref{tab:evidence-chain} says otherwise.  An intervention can raise
or lower every candidate at once, so subtracting the reference candidate's
change removes what is common to all of them.  Writing \(B_i\), \(B_j\),
\(B_k\) and \(A\) for the key, position, reference and donor-answer
candidates, garden, kitchen, hall and office in the example, the key effect
\(\kappa\), the position effect \(\pi\) and the donor-answer effect
\(\alpha\) are
\begin{equation}
 \kappa=\Delta\ell(B_i)-\Delta\ell(B_k),\qquad
 \pi=\Delta\ell(B_j)-\Delta\ell(B_k),\qquad
 \alpha=\Delta\ell(A)-\Delta\ell(B_k).
 \label{eq:contrasts}
\end{equation}

\paragraph{Tasks and models.}
The generators produce 17-fact, two-hop problems in three forms.  The
\emph{arrow} task uses arbitrary keys and numbers and the \emph{binding}
task names and apartment numbers; both carry the depth profiles and the
targeted interventions.  The \emph{word} task uses objects, owners and
places, and carries the memory experiments of Section~\ref{sec:suffix}.
The depth panel covers twelve models: sequential hybrids, which alternate
attention and recurrent layers; parallel hybrids, which combine them within
a layer; and dense transformers.  Targeted experiments use Nemotron-H,
Granite-H, RecurrentGemma-9B and Nano.  Every model runs as a raw
completion model and must answer 60\% of donor and recipient prompts before
it is assayed.  Section~\ref{sec:suffix} runs its memory swap on seven
hybrids, three of which do not read and count as part of its result;
Granite-H is not among them, since it does not reach that gate in the
memory roles (Appendix~\ref{app:memory}).  Appendix~\ref{app:setup} lists
the models, the identity pools and their roles.

\paragraph{Relation to prior work.}
\citet{feng2024how} identify binding representations, and
\citet{gur-arieh2026mixing} separate positional, lexical and reflexive
retrieval with counterfactual pairs.  We add a reference candidate and
independent variation of key and position, to follow a \emph{computed}
key.  For two-hop queries answered from parametric memory,
\citet{yang-etal-2024-large-language-models} and \citet{biran-etal-2024-hopping} locate where the
bridge entity is resolved and where the answer forms; our profiles measure
both for a key computed from the context, one transplant giving them at
every layer.  Attention-query transfer reveals reusable filtering
operations \citep{sharma2026llms}.  Work on hybrids removes a component,
zeroing an output or dropping one of the two caches
\citep{michalak2025some,afendulev2026attentionrecallsrecurrencecontrols};
\citet{zani2025contextual} find that a small subset of a hybrid's attention
heads suffices for retrieval, which Section~\ref{sec:models} reaches from
the other side, by measuring what a transplanted key does.  Removing a
component leaves the model in
a state it does not otherwise produce
\citep{makelov2024is,grant2026addressing}; we substitute instead, always an
activation the same model produced on a matched prompt, and
Appendix~\ref{app:alternatives} tracks how far the intervened trajectories
fall from natural ones.

\section{From key availability to answer resolution}
\label{sec:profile}

Using the pairs and scores of Section~\ref{sec:setup}, one transplant run
per layer input measures what the remaining computation does with each
state.  The layer across which \(\alpha\) rises most is the \emph{resolving
layer} (Section~\ref{sec:profile-steps}).  In Nemotron-H's panel of
Figure~\ref{fig:profile} the probe (green) first reads the key at L29, the
key effect (orange) builds to the input of L40, and across L40 it collapses
while the donor-answer effect (dotted) jumps.

\subsection{An intermediate key becomes available before the answer}


\begin{figure}[t]
  \centering
  \setlength{\abovecaptionskip}{-5pt}
  \includegraphics[width=0.9\linewidth]{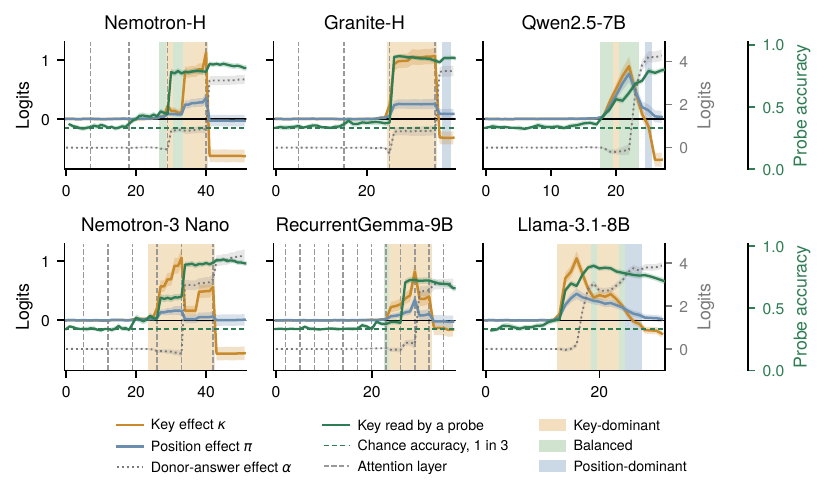}
  \caption{\textbf{Key availability rises before the main answer-conversion
  step.}  Transplants at every layer input in four hybrids and two dense
  transformers; 200 identities with 97.5\% identity-bootstrap intervals.
  The probe chooses among the donor identity's three keys, so its chance
  level is one in three.  Dense panels carry no attention markers because
  every one of their layers has attention.  The resolving layer's step
  makes the key effect negative in the three hybrids with four attention
  layers (Section~\ref{sec:profile-steps}; Appendix
  Table~\ref{tab:rise-window}).}
  \label{fig:profile}
\end{figure}

\begin{wraptable}[18]{R}{0.50\linewidth}   
\vspace{-14pt}
\centering
\footnotesize
\setlength{\tabcolsep}{3pt}
\begin{tabular}{lccc}
\toprule
 & Nemotron-H & Granite-H & Nano \\
\midrule
\multicolumn{4}{l}{\emph{Top-1 among seven (\%), chance 14}} \\
\quad rebinding & 86 & 71 & 67 \\
\quad row mismatch & 39 & 35 & 28 \\
\addlinespace[2pt]
\multicolumn{4}{l}{\emph{Causal scrubbing: donor shares only \dots{} (\(\times\))}} \\
\quad the key & 3.42 & 3.18 & 3.29 \\
\quad its own answer & 0.62 & 0.62 & 0.61 \\
\quad its row & 0.47 & 0.51 & 0.52 \\
\bottomrule
\end{tabular}
\caption{\textbf{The transferred state carries a key, and a row pointer with it.}  The three tests of
this section at the resolving layer's input
(Section~\ref{sec:profile-steps}), 200 identities each.  Top-1 is
how often the key's place wins among seven candidates; the scrubbing rows
give the factor by which the key's answer gains over the recipient's other
candidates, \(\times 1\) being no change.  Constructions, intervals, and
the preregistered form of the scrub test are in
Appendix~\ref{app:crossed}. }
\label{tab:acts-as-key}
\vspace{-8pt}
\end{wraptable}
A linear ridge probe reads the key out of the same states well before the
transplant converts it (Figure~\ref{fig:profile}, green;
Appendix~\ref{app:profile}), and in all twelve models its largest rise
falls across a layer that contains attention.  Only the five sequential
hybrids make that informative, since the others have attention in every
layer.  In those five attention sits in four to twelve of the layers, yet
the rise lands on an attention layer in all five: picking a layer at random
in each would do that about once in forty thousand, or once in three
thousand counting the two Nemotron-H checkpoints, which place attention
identically, as one.  What
happens between there and the layer that converts the key is the subject of
Section~\ref{sec:mechanism}.

In Nemotron-H and Granite-H the key effect rises to a \emph{key-dominant}
state at the input to the final attention layer (L40 and L35), \(\kappa\)
positive and reliably larger than \(\pi\), and the donor-answer effect
takes over after it (Figure~\ref{fig:profile}).  At peak key availability
Qwen2.5's state is \emph{balanced}, both effects supported and their
difference inside a prespecified band; its conversion spans several
attention layers and Nano's has two stages
(Section~\ref{sec:profile-steps}).  Appendix~\ref{app:profile} gives every
layer's effects and class, Appendix~\ref{app:sensitivity} the rules and the
constants fixed in advance.

\paragraph{The transferred state carries a key, and a row pointer with it.}
Three tests find the key dominant and the row pointer supported
(Table~\ref{tab:acts-as-key}).  In \emph{rebinding}, moving Alice to a
different place in the recipient while holding the donor state fixed moves
the promoted answer with her.  In \emph{row mismatch}, separating the key
from its row as in Section~\ref{sec:setup} leaves the effect following the
key.  In causal scrubbing \citep{chan2022causal}, resampling donor states
that share only the key keeps its answer ahead of the recipient's other
candidates; sharing only the answer or only the row reverses that
advantage.  Appendix~\ref{app:crossed} gives each test's construction and
its complete values.

\subsection{Peak key availability marks the largest conversion step}
\label{sec:profile-steps}

Write \(\alpha_l\) for the donor-answer effect of
Equation~\ref{eq:contrasts} at layer input \(l\), office against hall.
Scoring against the reference candidate keeps conversion apart from mere
suppression of the key candidate.  The increment
\(\Delta_l=\alpha_{l+1}-\alpha_l\) compares separate transplants made
immediately before and after layer \(l\).  The layer with the largest
increment is the \emph{resolving layer}; its input carries the peak key
effect in every profile we measured (Appendix
Table~\ref{tab:peak-bootstrap}), the landmark the later sections use.  It
opens the band of layers carrying the rise in six models, falls inside it
in three and closes it in three, so the alignment holds wherever in the
band it sits (Appendix Table~\ref{tab:rise-window}).  In Nemotron-H the
conversion begins at L29 and the key still builds to L40.

\paragraph{The step consumes the key.}
Across the resolving layer the key effect falls in all 23 profiles while
the donor-answer effect rises.  We call that decline consumption, in an
operational sense: a linear probe still reads the key out of the state
after the fall in all twelve models (Figure~\ref{fig:profile}, green;
Appendix~\ref{app:profile}).  How much falls follows what comes after the
resolving layer, since the score asks what the remaining computation can
still do with the key: the fall is complete in the three hybrids whose
resolving layer is their last attention layer, and partial wherever
attention follows, down to a quarter in Llama-3.1-8B (Appendix
Table~\ref{tab:rise-window}).

\paragraph{Key availability can rebuild between readers.}
Nano's penultimate (L33) and final (L42) attention layers both convert the
key; we call them its \emph{readers}.  On both tasks the key effect falls
across L33, the resolving layer, rebuilds mainly at recurrent layer L37,
and falls again across L42, in a second conversion that is smaller and
starts below the profile's peak (Figure~\ref{fig:profile}, bottom left;
Appendix Table~\ref{tab:panel}).  The landmark therefore marks the largest
conversion, and a key that a reader has already consumed can be resolved
again.
Section~\ref{sec:nano-later} tests whether L37 carries the key L42 converts.

\subsection{Attention carries the conversion}
\label{sec:concentration}

Is the rise in the donor-answer effect concentrated in one attention layer
or spread over many?  Let \(\mathcal T\) be the attention layers (every
layer in dense and parallel models) and take each one's share of the summed
positive increments, a negative increment getting no weight:
\begin{equation}
 p_l=\frac{\max(\Delta_l,0)}
 {\sum_{j\in\mathcal T}\max(\Delta_j,0)},\qquad
 C_{\rm attn}=\max_{l\in\mathcal T}p_l,\qquad
 e^H=\exp\left(-\sum_{l\in\mathcal T}p_l\log p_l\right).
 \label{eq:increment}
\end{equation}
Here \(C_{\rm attn}\) is the largest share and \(e^H\), the exponential of
the shares' Shannon entropy, is the effective attention count, as perplexity
is an effective vocabulary size \citep{hill1973diversity,jost2006entropy}:
two equal contributors give two, one gives one.

\begin{figure}[t]
 \centering
 \setlength{\abovecaptionskip}{-5pt}
 \includegraphics[width=\linewidth]{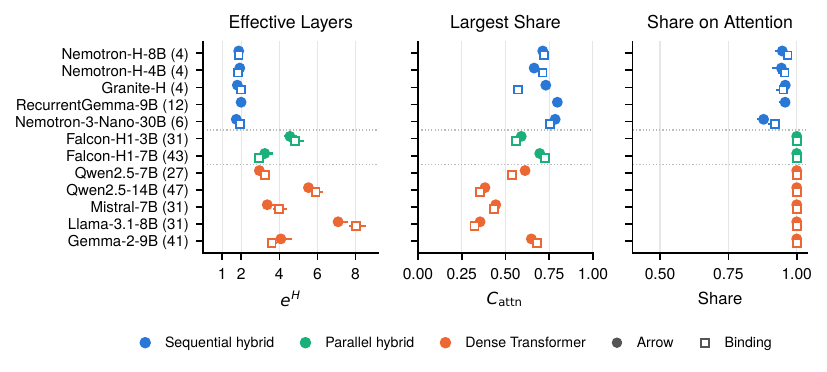}
 \caption{\textbf{The conversion uses the attention available in a late
 band of depth.}  Each row is a model, filled circles the arrow task and
 hollow squares the binding task, with 95\% identity-bootstrap intervals.
 Left: the effective attention count carrying the rise,
 \(e^H\).  Middle: the share of it carried by the single largest attention
 layer, \(C_{\rm attn}\).  Right: the share falling on attention layers at
 all, which only the sequential hybrids make informative.  Appendix
 Table~\ref{tab:concentration-null} gives the first two against a uniform
 allocation.}
 \label{fig:concentration}
\end{figure}

Across twelve models and two tasks the rise occupies a similar late band of
depth, and inside it the conversion uses the attention available.  Every
sequential hybrid has two attention layers in the band and spreads the rise
over \textbf{1.7--2.0} of them; the models with attention in every layer
spread it over \textbf{2.9--8.0}, and even RecurrentGemma, with twelve
attention layers, confines the rise to the two in the band
(Figure~\ref{fig:concentration}; Appendix Tables~\ref{tab:panel}
and~\ref{tab:rise-window}).  Against a uniform allocation over each model's
own attention layers the ordering reverses, since the dense models use a
small part of the attention they have and the hybrids nearly all of theirs,
so the effective attention count reports where the attention sits rather
than how concentrated a model is
(Appendix Table~\ref{tab:concentration-null}).

What availability does not explain is who does the converting.  Attention
carries 95\% of Nemotron-H's rise while occupying four of the 28
token-mixing layers among its 52,
and no sequential hybrid leaves recurrent layers more than an eighth of it
(Appendix Table~\ref{tab:panel}).  How much recurrence takes depends on the
task: on the word task, where these layers can read a binding from memory
(Section~\ref{sec:suffix}), attention's share in Nemotron-H falls to 0.86,
and it returns on the numeric twin, where the swap finds no read (Appendix
Table~\ref{tab:word-share}).  Most of the conversion stays in attention on
both tasks.  Intervening at the final token does not decide this between
the two: a recurrent layer after the intervened site combines the injected
residual with a memory built from the recipient's own unmodified prefix
(Equation~\ref{eq:ssd}), so it has the same opportunity to resolve the key
that attention has, and Section~\ref{sec:suffix} shows it takes that
opportunity where the task allows.  These comparisons span checkpoints
trained differently; Section~\ref{sec:mechanism} shows that attention heads
carry the lookup within each hybrid.

\section{How recurrence and attention implement the handoff}
\label{sec:mechanism}

The profiles locate a usable key and the layer where it becomes an answer.
This section asks which components carry the key to the reader and which
read it.

\subsection{Recurrent outputs carry the key to the reader}
\label{sec:causal}

Two donors differ only in their key, Alice or Bob, keeping office as the
answer and the completing fact at the same row.  In the Alice run we swap
one recurrent layer's final-token output for Bob's, continue to the
resolving layer's input, and transplant that state into the recipient,
which puts Alice in the garden and Bob in the kitchen; if that output
carries the key the lookup should move from garden to kitchen.  Write
\(\kappa_{\rm Bob}\) and \(\kappa_{\rm Alice}\) for the key effect of
Section~\ref{sec:setup} read for Bob's place and for Alice's, after
transplanting either donor's own state or the swapped one.  We report
both:
\begin{equation}
 g=\frac{\kappa_{\rm Bob}(\text{swap})}{\kappa_{\rm Bob}(\text{Bob})},
 \qquad
 h=\frac{\kappa_{\rm Alice}(\text{swap})}{\kappa_{\rm Alice}(\text{Alice})},
 \label{eq:writer-swap}
\end{equation}
Bob's lookup \(g\), which Figure~\ref{fig:writer-swap} plots, and Alice's
lookup \(h\) (the guest and host lookups of Appendix~\ref{app:writers}); an
unchanged key gives \(g=0\) and \(h=1\).  A separate depth profile ranks the
recurrent layers before the reader by how much each raises the key effect.
We swap the highest-ranked, the \emph{nominated} layers, and as many nearby
recurrent layers whose contribution there is near zero, the \emph{matched}
layers (Appendix~\ref{app:writers}).

Swapping the nominated outputs switches part of the lookup to Bob in all
four hybrids and reduces Alice's (Figure~\ref{fig:writer-swap}, left and
middle).  The two halves do not travel together.  Matched layers, chosen
because their contribution to the key is near zero, create some of Bob's
lookup as well, and in Nemotron-H almost as much as the nominated layers
do; what they do not do is remove Alice's, which they leave nearly intact
where the nominated layers take it down (Appendix
Table~\ref{tab:writer-swap}).  So the nominated outputs are what the reader
needs to find Alice, and part of what it needs to find Bob.  As a control
we swap between two donors that share the key but differ in the answer; the
lookup then stays where it was, so the removal follows the key rather than
the damage.  How the key is carried differs by model: Granite-H relies on
one nominated recurrent layer while its other two move neither lookup, and
Nemotron-H and Nano spread it over several.  The nominated layers straddle the
attention layer where the probe first reads the key, so the swap shows that
these outputs carry key-specific information into the reader without saying
which of them prepared it.

\begin{figure}[t]
  \centering
  \setlength{\abovecaptionskip}{-0pt}
  \includegraphics[width=\linewidth]{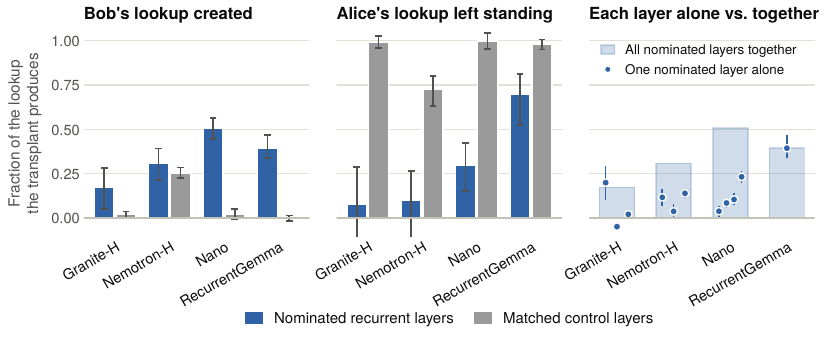}
  \caption{\textbf{Recurrent outputs carry the key to the reader.}  Both
  lookups after the swap, each as a fraction of the lookup its own
  transplant produces, with 95\% identity-bootstrap intervals; \(g\) and
  \(h\) of Equation~\ref{eq:writer-swap}.  Left: Bob's lookup created, all
  nominated layers against all matched controls.  Middle: Alice's lookup
  left standing, the same two sets, which is where they separate.  Right:
  each nominated layer alone (points) against the joint swap (bar), which
  can fall below zero when one layer moves the lookup toward Alice.
  Appendix Table~\ref{tab:writer-swap} reports both beside the
  redirection.}
  \label{fig:writer-swap}
\end{figure}

\subsection{A few attention heads read the key}
\label{sec:models}
\label{sec:site}
\label{sec:route}

\begin{wraptable}{R}{0.53\linewidth}
\vspace{-14pt}
\centering
\footnotesize
\setlength{\tabcolsep}{3pt}
\begin{tabular}{lccc}
\toprule
 & 8-head & All heads & Random sets \\
 & injection & clamped & beaten \\
\midrule
Nemotron-H (L40) & \(\approx\)100\% & \(\approx\)0\% & 100\% \\
Granite-H (L35) & 69--75\% & \(\approx\)0\% & 100\% \\
Nano (L33) & 93--96\% & 36--39\% & 100\% \\
\bottomrule
\end{tabular}
\caption{\textbf{A few heads read the key.}  Columns give the share of the
transplant's key effect reproduced by injecting the eight heads' changes,
the share surviving a clamp of all the layer's heads, and the fraction of
random eight-head sets beaten; ranges span rebinding and row mismatch
(Appendix Tables~\ref{tab:route}--\ref{tab:route-nano}).}
\label{tab:heads-read}
\vspace{-5pt}
\end{wraptable}
At the resolving layers of Nemotron-H, Granite-H and Nano
(Table~\ref{tab:heads-read}) we intervene on attention-head outputs, on the
data of Section~\ref{sec:setup}, recording how each head's output changes
under the transplant.  \emph{Injecting} a set's changes into an otherwise
unmodified recipient asks whether those heads alone make garden rise;
\emph{clamping} all the layer's heads to their unmodified outputs asks what
is left without them.

Eight heads, chosen by activation patching on separate examples
(Appendix~\ref{app:reader}), carry most of the transplant's effect and beat
every random set of eight they were tested against; Granite-H spreads the
reading over more heads.  Clamping all heads removes the key effect
wherever no later reader follows (Table~\ref{tab:heads-read}).  Giving the
heads their donor-prompt outputs instead removes most of the key effect:
they compute the answer from the recipient's facts rather than relay what
they computed in the donor.

\subsection{Recurrence can carry the key again between readers}
\label{sec:nano-later}

Nano's key effect falls across its first reader (L33) and rebuilds before
its final one (L42), most of it at the recurrent layer L37
(Figure~\ref{fig:profile}).  L37's contribution is hidden while the first
reader still has the key, so we remove the key before L33: a second run
asks about an object whose owner has no fact of their own, and its
final-token outputs at the recurrent layers before L33 replace the host's.
The prompt's facts stay as they are, so the layers after L33 can still find
Alice, and a lookup that depends on the key survives at the final reader.
Replacing L37's output as well removes 64\% of it; the same replacement at
a comparison layer in the same span removes none; and putting L37's
undisturbed output back recovers 57\% of what was lost.  So L37 supplies
part of what the final reader uses, and the reader's heads matter more once
the early key is weakened (Appendix Table~\ref{tab:late-supply} and
Figure~\ref{fig:delayed-key}).

\section{Recurrent memory can retrieve after the last attention layer}
\label{sec:suffix}

In the hybrids studied here, two to eight recurrent layers follow the last
attention layer and can no longer call it (Appendix
Table~\ref{tab:memory-map}).  Each keeps a memory \(m_l\) of the preceding
tokens (Section~\ref{sec:setup}), which in their Mamba-2 layers is a
decayed sum of outer products.  Dropping the layer index, the memory after
token \(t\) and its reading are
\begin{equation}
 m_t=a_t m_{t-1}+B_t x_t^{\top},
 \qquad
 y_t=m_t^{\top}C_t=\sum_{s\le t}\Bigl(\prod_{r=s+1}^{t}a_r\Bigr)
 \bigl(C_t^{\top}B_s\bigr)\,x_s,
 \label{eq:ssd}
\end{equation}
\citep{dao2024transformersssmsgeneralizedmodels}: each earlier token writes
a value \(x_s\) under a key \(B_s\), the decay factors \(a_r\) weight the
earlier writes, and the final token retrieves the values whose keys match
the query \(C_t\) it computes, an associative memory in the fast-weight
sense \citep{schlag2021linear}.  Replacing \(m_l\) changes what is stored
under which key and leaves that query untouched, so a hybrid in which these
layers can look up a binding has a second retrieval route.  We test it by
exchanging two people's places in the word task, which its facts allow
without adding or removing a word.  Earlier interventions on Nemotron-H's
post-attention Mamba outputs motivate the test
(Appendix~\ref{app:suffix-retrieval}).

\begin{wraptable}{R}{0.5\linewidth}
\vspace{-12pt}
\centering
\footnotesize
\setlength{\tabcolsep}{3pt}
\begin{tabular}{lccc}
\toprule
 & Test & Control & Transposed \\
\midrule
Alice & Garden & Garden & Kitchen \\
Bob & Kitchen & Kitchen & Garden \\
Carol & Hall & Hall & Hall \\
\midrule
Question & \shortstack{Where is\\the ball?} & \shortstack{Where is\\the cup?} & \shortstack{Where is\\the cup?} \\
Own answer & Garden & Hall & Hall \\
\bottomrule
\end{tabular}
\vspace{-8pt}
\end{wraptable}
\paragraph{Writing a different fact into memory.}
The test prompt is the story of Section~\ref{sec:setup}: the ball belongs
to Alice, and Alice is in the garden.  Let \(S\) be the recurrent layers
after the last attention layer.  We run the test prompt twice and, at its
final token, install in every \(l\in S\) the memories of a separate
\emph{memory prompt}, \(m_l\leftarrow m_l(x^{\rm ctrl})\) in one run and
\(m_l\leftarrow m_l(x^{\rm trans})\) in the other.  Everything else comes
from the test prompt: its tokens, its question, and the residual entering
\(S\).  Both memory prompts ask about Carol's cup and answer hall; they
differ only in exchanging Alice's and Bob's places, so the transposed one
puts Alice in the kitchen.  Reading Alice's binding predicts kitchen and
copying the memory prompt's own answer predicts hall, and we require
kitchen to gain on both hall and garden, the test prompt's own answer.  For
each candidate place \(c\), \(\Delta\ell(c)\) is its transposed-minus-control
logit change, which cancels what is common to both:
\begin{equation}
 \sigma_{\rm own}=\Delta\ell(\text{kitchen})-\Delta\ell(\text{hall}),\qquad
 \sigma_{\rm test}=\Delta\ell(\text{kitchen})-\Delta\ell(\text{garden}).
 \label{eq:memory-read}
\end{equation}
Exponentiating a mean score gives the geometric-mean factor by which the
exchange multiplies one candidate's probability relative to the other
(Appendix~\ref{app:memory}).  Under \emph{ordinary} execution the last
attention layer has already put the test prompt's own answer several logits
ahead, so a read of the memory can move the odds without deciding the
answer.  To ask what the read does when that lead is not already there, we
run the swap a second time with the last attention layer's final-token
update left out, the \emph{skipped} state.  The two differ only in the
residual entering \(S\) at the final token.

\begin{figure}[t]
 \centering
 \setlength{\abovecaptionskip}{-0pt}
 \includegraphics[width=\linewidth]{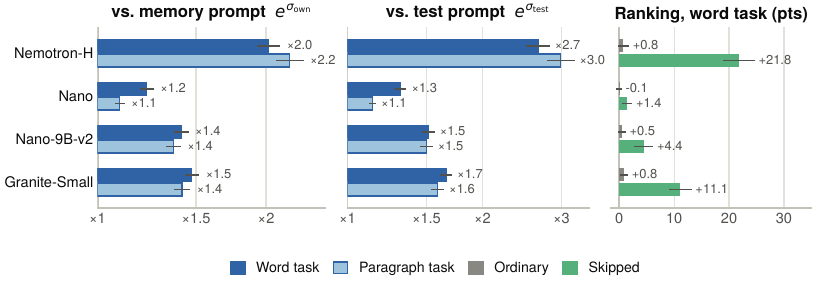}
 \caption{\textbf{Changing a stored location moves the answer in four
 hybrids.}  Left and middle: geometric-mean factors for kitchen's
 probability against hall (\(e^{\sigma_{\rm own}}\)) and against garden
 (\(e^{\sigma_{\rm test}}\)), on logarithmic axes starting at
 \(\times1\).  Both are read under ordinary execution; shade gives the
 task version, word or paragraph.  Right: the change in how often the
 memory's place ranks first, on the word task, with color giving the
 incoming state, ordinary or skipped.
 Intervals are 95\% over 200 identities; Appendix Table~\ref{tab:memory-map} lists every model and
 version tested.}
 \label{fig:memory-read}
\end{figure}

\paragraph{The recurrent layers read the binding in four hybrids.}
The exchange promotes kitchen in all four models
(Figure~\ref{fig:memory-read}).  In Nemotron-H it raises the
kitchen-to-garden probability ratio by a geometric-mean factor of
\textbf{2.7}, in every one of the 200 identities; the other three rise by
smaller factors, each in at least 91\% of identities taken one at a time.
Ordinary execution changes the odds without changing the ranking.  Skipping
the last attention layer's final-token update removes the test answer's
lead, and the read against that answer grows rather than weakening: the
memory's place takes the top-1 answer from the test prompt's in
\textbf{22} percentage points more identities in Nemotron-H,
\textbf{11} in Granite-4.0-H-Small, \textbf{4.4} in Nemotron-Nano-9B-v2 and
\textbf{1.4} in Nano (Figure~\ref{fig:memory-read}, right), so the read
does not run through the last attention layer.  Each model repeats the result on 200 new
identities with the facts written as one paragraph, and Appendix
Table~\ref{tab:memory-map} adds a second sentence template and a version
without the task's three distractor facts.

\paragraph{Controls.}
The control memory changes only the question the memory was built under,
and installing it does not raise the memory prompt's own answer, hall, in
three of the four models; it raises it slightly in the fourth, whose scores
are net of that effect.  The read is also specific to the binding the
question asks about: in Nemotron-H, exchanging two places the question does
not ask about moves the queried place's odds by \(\times 1.06\), against
\(\times 2.7\) when the exchange reaches that binding, and Nano-9B-v2 and
Granite-Small give the same picture.  The answers also
depend on the post-attention layers themselves (Appendix~\ref{app:memory}
and~\ref{app:dependence}).  Recurrent layers after the last attention layer
can thus retrieve a binding from their own memory.

\begin{wraptable}[10]{R}{0.52\linewidth}   
\vspace{-12pt}                              
\centering
\footnotesize
\setlength{\tabcolsep}{4pt}
\begin{tabular}{lcc}
\toprule
 & Nemotron-H & Granite-Small \\
\midrule
Key over row & \(\times\)\,1.9 & \(\times\)\,1.4 \\
Key over test prompt's & \(\times\)\,2.4 & \(\times\)\,1.5 \\
Row over test prompt's & \(\times\)\,1.3 & \(\times\)\,1.1 \\
\bottomrule
\end{tabular}

\caption{\textbf{The answer follows the key.}  Geometric-mean factors after
rotating the key rows, in ordinary execution; 200 identities per model.
Intervals, the share of identities, and the construction are in Appendix
Table~\ref{tab:crossed-appendix}.}
\label{tab:crossed-memory}
\vspace{-6pt}
\end{wraptable}
\paragraph{The read is addressed by the key, not by the row.}
With the test prompt's row order the two accounts predict the same place,
since what the memory binds to the queried key also sits in that key's
row.  We separate them by rotating the key rows while keeping the
bindings: the place bound to the queried key moves into another key's row,
and the row that key occupies holds a third place.  The answer follows the
key in both models (Table~\ref{tab:crossed-memory}), so the key addresses the content.

\paragraph{Which models use it.}
Three kinds of null separate architecture, model and task.
RecurrentGemma-9B keeps a vector state updated element by element, with no
outer product for a query to address, so a key-addressed read has no direct
counterpart there, though its output gate can still select per channel.
Nemotron-H-47B and Nano-12B-v2 have Mamba-2 memories and show no read even
so, while passing the same exact checks as the positive runs.  Nano-9B-v2,
pruned and distilled from Nano-12B-v2 \citep{nvidia2025nemotronnano2}, reads
where its parent does not: neither the pretraining corpus nor the
architecture family settles which models use it.  Nemotron-H also shows no
read on the numeric version of the task, where the post-attention layers
take a third as much of the answer conversion
(Section~\ref{sec:concentration}), so reading and conversion vary together.
Single-token room numbers leave the read where the multi-token version left
it in Nemotron-H and Granite-Small (Appendix~\ref{app:memory}), so length is
not what the read needs of a value.

\section{Discussion}
\label{sec:families}
\label{sec:robustness}
\label{sec:robust}

A computed key moves through a sequential hybrid as a handoff.  Earlier work showed that retrieval in hybrids depends on attention: zeroing
attention-head outputs removes it, with no recurrent compensation
\citep{michalak2025some}, and pretrained hybrids delegate the aggregation
step of in-context retrieval to their attention layers
\citep{bick2025understanding}; in recurrent models associative recall is
carried by components such as Mamba's short convolutions not by the
recurrence itself \citep{arora2025mechanistic}.  Our assay places these
findings in one computation, locating where the key becomes usable and
where it becomes the answer, and which components supply it and which read it. The depth profile agrees with the sites an activation-patching screen chose
(Appendix~\ref{app:reader}).  Where the read appears, the recurrent state
behaves like the fast weights
of \citet{schlag2021linear}: written while the prompt is read, holding
bindings under the keys that address them, and rewritable, since
installing the memory of a prompt whose facts are exchanged moves the
answer.  What addresses it is the query the model computes, not a token in
the prompt: the question names neither the person nor the place.

\paragraph{Scope.}
The tasks are generated fact lists read in raw completion mode, and each
targeted intervention covers three to four hybrids; the assay needs only
prompt pairs and a transplant, so natural text and other architectures are
a direct extension rather than a tested one.  The twelve checkpoints differ
in training as well as in architecture, so the panel describes where the
conversion happens and not what makes models differ.  How far the reading
concentrates, and whether the memory read reaches the model's first choice,
are themselves model-dependent (Table~\ref{tab:heads-read};
Figure~\ref{fig:memory-read}).  What a value must be for these layers to
read it is open too: the read is strong on place words and near zero on room numbers, and neither single-token numbers nor number words recover it (Appendix~\ref{app:memory}). Across these experiments a key becomes an answer through a handoff
between recurrence and attention, and in four of the seven hybrids
retrieval can continue past the last attention layer.

\raggedbottom  
\newpage

\section*{Reproducibility statement}
Every experiment runs from a configuration file that fixes the model and
its revision, the intervention site, the identity pool and its seed, and
the numerical thresholds, all written before the held-out sample was run;
Appendix Table~\ref{tab:evidence-chain} lists each experiment with its
data and its inferential role, and the supplement carries the
configurations, their SHA-256 hashes, and the serialized prompts.  The
identity pools are generated from a seed and a tokenizer, so they can be
rebuilt exactly.  Appendix~\ref{app:setup} gives the models, the accuracy
gate every model passes before it is assayed, and the exact checks each run
reports.

\section*{Use of large language models}
The authors used a large language model as a coding and writing assistant:
it helped implement and run analysis scripts, draft and revise text, and
check the paper against its own tables.  Every experiment, result and claim
was designed, verified and approved by the authors, who take full
responsibility for the contents.

\newpage
\bibliography{references}
\bibliographystyle{iclr2027_conference}

\newpage
\appendix

\FloatBarrier
\section{Experimental setup and notation}
\label{app:setup}
\label{app:inventory}
\label{app:common}

\paragraph{Terminology.}  The paper and its tables use one vocabulary; the
released runs and configuration files keep the field names the assay was
built with, where the key is called \emph{content} and the reference
candidate the \emph{neither} candidate, as in
\texttt{content\_vs\_neither} and the scrub \(S_C\).

\begin{table}[htbp]
\centering
\scriptsize
\setlength{\tabcolsep}{2pt}
\begin{tabular}{
  >{\raggedright\arraybackslash}p{0.20\linewidth}
  >{\raggedright\arraybackslash}p{0.24\linewidth}
  >{\raggedright\arraybackslash}p{0.28\linewidth}
  >{\raggedright\arraybackslash}p{0.18\linewidth}}
\toprule
Stage & Intervention & Data & Role \\
\midrule
Flow localization & Residual patch & 20 paired & Exploratory \\
Head ranking & Head-output patch & 30 discovery + 300 held-out & Confirmatory \\
Fixed-donor crossing & Whole-state transplant & 200 graphs / 600 crosses & Confirmatory \\
Fixed-recipient crossing & Whole-state transplant & 200 graphs / 600 crosses & Confirmatory \\
Common assay: factorial & Whole-state transplant & 200 identities / 1,800 cells & Confirmatory \\
Common assay: path patching & Head-delta injection, clamps & Same 200 identities & Confirmatory \\
Common assay: scrubbing & Matched resampling & Same 200 identities & Confirmatory \\
Cross-model panel & Prespecified screen + held-out run & Six checkpoints & Confirmatory \\
Granite-H factorial & Whole-state transplant, pre-L35 & Same 200 identities & Confirmatory \\
Granite-H route and scrubs & Head-delta injection, clamps, causal scrubbing & Same 200 identities; 30 discovery & Confirmatory \\
Depth profile & Whole-state transplant at every layer & Same 200 identities, three models & Descriptive \\
Second prompt format & Whole-state transplant, pre-L39/L40 & 200 identities, second-format pool & Confirmatory \\
Probe reference & Ridge probes at every layer & 240/80/200 identities & Descriptive \\
Panel profiles & Whole-state transplant at every layer & Same 200 identities, seven further checkpoints & Confirmatory (thresholds) \\
Shifted transplant & Whole-state transplant one layer later & Same 200 identities, three models & Confirmatory \\
Binding task & Whole-state transplant at every layer & 200 identities, binding pool, full panel & Confirmatory (class; thresholds) \\
Peak alignment & Compare key peak and largest answer-conversion increment & 23 profiles across twelve models & Descriptive; repeated on Nano \\
Writer swap & Recurrent-output replacement before attention & 200 identities, four hybrids & Confirmatory \\
Suffix disruption and rescue & Mamba-output clamp and cross-donor rescue & 200 identities in each of two word pools & Confirmatory \\
Recurrent-memory read & Recurrent-state replacement after attention & 200 fresh identities per model and wording & Confirmatory \\
Memory-swap trajectory audits & Recipient-reference tail rates, wording-matched & Same 200 identities per template; 800-identity references & Descriptive \\
Output dependence & Final-token Mamba-output replacement, no transplant & 100 fresh identities; 100-identity calibration & Confirmatory \\
\bottomrule
\end{tabular}
\caption{Experiments, data, and inferential roles.}
\label{tab:evidence-chain}
\end{table}

\paragraph{Input matrices and the running example.}
Each prompt renders 17 facts as an ordered list, so a row index is prompt
order, not graph distance.  Figure~\ref{fig:handoff}b renames one saved
assay pair (cell \texttt{mvp17-test-0000-oa-c0-a1-p1}) with names and
places for readability; the figure source records the renaming, and the
matrix below shows the pair.
\begin{center}
\begin{tabular}{cc}
\textbf{Donor facts} & \textbf{Recipient facts} \\
$\begin{array}{r}1\\2\\3\\4\\5\\6\\7\\8\\9\\10\\11\\12\\13\\14\\15\\16\\17\end{array}\;\left[\begin{array}{ll}\color{black}\texttt{UBV}&\color{black}32\\\color{black}\texttt{QNV}&\color{black}\texttt{JTJ}\\\color{black}\texttt{EUG}&\color{black}36\\\color{black}\texttt{CEZ}&\color{black}\texttt{WNV}\\\color{black}\texttt{VFS}&\color{black}31\\\color{black}\texttt{IMS}&\color{black}\texttt{YED}\\\color{black}\texttt{BDH}&\color{black}\texttt{JCF}\\\color{black}\texttt{MJH}&\color{black}\texttt{RAY}\\\color{black}\texttt{TUV}&\color{black}38\\\color{black}\texttt{GWY}&\color{black}\texttt{TUV}\\\color{black}\texttt{RAY}&\color{black}35\\\color{black}\texttt{SBW}&\color{black}34\\\color{black}\texttt{KOD}&\color{black}\texttt{EUG}\\\color{blue!65!black}\texttt{CFG}&\color{blue!65!black}39\\\color{black}\texttt{XXS}&\color{black}\texttt{OVJ}\\\color{black}\texttt{WNV}&\color{black}\texttt{BDH}\\\color{black}\texttt{NOG}&\color{black}\texttt{CFG}\end{array}\right]$ & $\begin{array}{r}1\\2\\3\\4\\5\\6\\7\\8\\9\\10\\11\\12\\13\\14\\15\\16\\17\end{array}\;\left[\begin{array}{ll}\color{black}\texttt{RAY}&\color{black}35\\\color{black}\texttt{WNV}&\color{black}\texttt{BDH}\\\color{orange!80!black}\texttt{CFG}&\color{orange!80!black}32\\\color{black}\texttt{NOG}&\color{black}\texttt{YED}\\\color{black}\texttt{UBV}&\color{black}39\\\color{black}\texttt{CEZ}&\color{black}\texttt{WNV}\\\color{black}\texttt{IMS}&\color{black}\texttt{CFG}\\\color{black}\texttt{GWY}&\color{black}\texttt{TUV}\\\color{black}\texttt{SBW}&\color{black}34\\\color{black}\texttt{QNV}&\color{black}\texttt{JTJ}\\\color{black}\texttt{EUG}&\color{black}36\\\color{black}\texttt{VFS}&\color{black}31\\\color{black}\texttt{BDH}&\color{black}\texttt{JCF}\\\color{blue!65!black}\texttt{TUV}&\color{blue!65!black}38\\\color{black}\texttt{XXS}&\color{black}\texttt{OVJ}\\\color{black}\texttt{KOD}&\color{black}\texttt{EUG}\\\color{black}\texttt{MJH}&\color{black}\texttt{RAY}\end{array}\right]$ \\
Query: \texttt{NOG}; answer: $39$ & Query: \texttt{MJH}; answer: $35$
\end{tabular}
\end{center}
Each row is rendered as \texttt{Key <key> -> Value <value>}.
Both prompts begin with the same three solved one-step demonstrations:
\texttt{MIH}$\to90$, \texttt{EAF}$\to70$, and \texttt{IHE}$\to40$.
The task question is \texttt{Starting from <query>, follow mappings until
reaching a number. Answer:}.  The candidate answers are
$\{31,32,34,35,36,38,39\}$.  Keys and candidate values are renamed solely
for presentation; all bindings and row positions are preserved.

\paragraph{Formal crossing and selectivity definitions.}
Every run is scored on the same seven candidate answers: three
donor-answer values (the donor variants of the factorial share them), the
three values the recipient binds to the keys under test, and the
recipient's own answer; top-1 accuracy is taken among these seven.
A candidate-level account \(H\) maps a donor--recipient pair to its
predicted answer.  A fully discriminating crossing satisfies
\begin{equation}
 H_i(x_d,x_r)\ne H_j(x_d,x_r)\quad\text{for all }i\ne j.
 \label{eq:crossing}
\end{equation}
For a predicted value \(B_j\) among three recipient-bound values,
\begin{equation}
 S(B_j)=\Delta\ell(B_j)-\tfrac12\sum_{k\ne j}\Delta\ell(B_k).
 \label{eq:selectivity}
\end{equation}
We call this mapping selectivity \(S^{\rm map}\) in the fixed-donor
crossing and key selectivity \(S^{\rm con}\) in the fixed-recipient
crossing.  Tracking chooses the largest logit increase among these three
values (chance 33\%); top-1 chooses the highest final logit among all seven.

\paragraph{Notation.}  Table~\ref{tab:notation} collects the symbols of the
main text.

\begin{table}[htbp]
\centering
\scriptsize
\setlength{\tabcolsep}{4pt}
\begin{tabular}{l>{\raggedright\arraybackslash}p{0.74\linewidth}}
\toprule
Symbol & Meaning \\
\midrule
\(x_d\), \(x_r\); \(h_l(x)\) & Donor and recipient prompts; the answer-position state entering layer \(l\) (``pre-L\(l\)'') \\
\(h_l\), \(o_l\), \(m_l\) & At the final token: the residual state entering layer \(l\); the output a recurrent layer adds to it (\(h_{l+1}=h_l+o_l\)); the recurrent layer's memory, accumulated over the preceding tokens.  Transplants replace \(h_l\), writer swaps \(o_l\), memory swaps \(m_l\) \\
\(\Delta\ell(c)\) & Change in the logit of candidate \(c\) when the donor's state replaces the recipient's (Eq.~\ref{eq:transplant}) \\
\(A\); \(B_1,B_2,B_3\); \(A_r\) & The donor's answer; the values the recipient binds to the three keys; the recipient's own answer \\
\(\mathcal{A}\), \(\mathcal{P}\), \(\mathcal{K}\) & The fixed-answer, position, and key causal models \\
\(S(B_j)\); \(S^{\rm map}\), \(S^{\rm con}\) & Selectivity of a candidate over the other two bindings; mapping (fixed-donor) and key (fixed-recipient) selectivity (Eq.~\ref{eq:selectivity}) \\
\(\kappa\), \(\pi\) & Key-vs-reference and position-vs-reference contrasts (Eq.~\ref{eq:contrasts}) \\
Tracking, top-1 & Rate at which the key-predicted candidate rises most among \(B_{1:3}\); rate at which it ranks first among all seven candidates \\
\(F\); \(P_{\rm T}\), \(P_{\rm all}\), \(P_{\bar{\rm T}}\), \(P_{\rm B}\) & Whole-state transplant; head-output injection for the top-eight, all, other, and bottom-eight heads (Eq.~\ref{eq:path}) \\
\(R_{\rm by}\), \(D_{\rm attn}\), \(C_{\rm T}\) & Residual bypass, donor-clamped heads, and top-eight heads clamped under \(F\) \\
\(\alpha_l\), \(\Delta_l\), \(C\) & The donor-answer effect (Eq.~\ref{eq:contrasts}), its per-layer increment (Eq.~\ref{eq:increment}), and its all-layer concentration (Appendix~\ref{app:sensitivity}) \\
\(S_C\), \(S_A\), \(S_P\) & Scrubs that preserve the key, the answer, or the row (Eq.~\ref{eq:scrub}) \\
\bottomrule
\end{tabular}
\caption{Notation.}
\label{tab:notation}
\end{table}

\paragraph{Models.}
The primary model is Nemotron-H-8B.  The crossed-assay panel was fixed
before any transplant, and the depth-profile panel adds further
checkpoints (Table~\ref{tab:model-panel}); models below the fixed accuracy
threshold are not analyzed further.

\begin{table}[htbp]
\centering
\scriptsize
\setlength{\tabcolsep}{3pt}
\resizebox{\textwidth}{!}{%
\begin{tabular}{llcclll}
\toprule
Checkpoint & Revision & Class & Layers & Task & Site & Outcome \\
\midrule
\texttt{nvidia/Nemotron-H-8B-Base-8K} & 94ea861e & H & 52 & Pass & pre-L40 & Primary model \\
\texttt{ibm-granite/granite-4.0-h-micro-base} & 372ede0b & H & 40 & Pass & pre-L35 & Both crossings pass \\
\texttt{ibm-granite/granite-4.0-h-small-base} & 4b4faa80 & H & 40 & Pass & --- & Memory swap only \\
\texttt{Qwen/Qwen2.5-7B} & d1497293 & D & 28 & Pass & pre-L22/26\(^\ddagger\) & Balanced / reference \\
\texttt{mistralai/Mistral-7B-v0.3} & caa1feb0 & D & 32 & Pass & pre-L19 & Profile panel \\
\texttt{Zyphra/Zamba2-2.7B} & 31afeeac & H & 54 & Fail\(^\S\) & --- & Below accuracy gate \\
\texttt{allenai/OLMo-2-1124-7B} & 7df9a825 & D & 32 & Fail & --- & Below accuracy gate \\
\texttt{tiiuae/Falcon-H1-7B-Base} & c9a4cbb9 & P & 44 & Pass & pre-L29 & Profile panel \\
\midrule
\texttt{nvidia/Nemotron-H-4B-Base-8K} & faba3b73 & H & 52 & Pass & pre-L40 & Profile only \\
\texttt{tiiuae/Falcon-H1-3B-Base} & c096902c & P & 32 & Pass & pre-L21 & Profile only \\
\texttt{ibm-ai-platform/Bamba-9B-v2} & b42852dc & H & 32 & Fail\(^\S\) & --- & Below accuracy gate \\
\texttt{Zyphra/Zamba2-7B} & 0ed988f3 & H & 81 & Fail\(^\S\) & --- & Below accuracy gate \\
\texttt{Qwen/Qwen2.5-14B} & 97e1e763 & D & 48 & Pass & pre-L36 & Profile only \\
\texttt{meta-llama/Llama-3.1-8B} & d04e592b & D & 32 & Pass & pre-L16 & Profile only \\
\texttt{google/gemma-2-9b} & 33c19302 & D & 42 & Pass\(^\S\) & pre-L28 & Profile only \\
\texttt{google/recurrentgemma-9b} & 141ed2b6 & H & 38 & Pass\(^\S\) (arrow) & pre-L29 & Third family; binding gate 55/53\% \\
\texttt{nvidia/NVIDIA-Nemotron-3-Nano-30B-A3B-BF16} & bf77c317 & H & 52 & Pass & pre-L33 & Profile and targeted experiments; post-trained release, run without its chat template \\
\midrule
\texttt{nvidia/NVIDIA-Nemotron-Nano-9B-v2-Base} & dc0661c8 & H & 56 & Pass & --- & Memory swap only \\
\texttt{nvidia/NVIDIA-Nemotron-Nano-12B-v2-Base} & 78dc93a7 & H & 62 & Pass & --- & Memory swap only \\
\texttt{nvidia/Nemotron-H-47B-Base-8K} & 81a3fb4f & H & 98 & Pass & --- & Memory swap only \\
\bottomrule
\end{tabular}}
\caption{Model panel.  H: sequential hybrid; P: parallel hybrid (attention
in every layer); D: dense Transformer.  ``Task: Fail'' means unmodified
accuracy fell below the fixed threshold; \(^\S\)evaluated with the model's
beginning-of-sequence token after failing without it (Bamba-9B-v2 and
both Zamba2 checkpoints fail both ways).  For the profile
panel the site column gives the layer with the largest key effect, and
Table~\ref{tab:panel} gives each model's concentration and effective
attention count.
Crossed-assay sites were chosen by the prespecified screen, and every prespecified site and threshold is in the frozen configuration files whose SHA-256 the supplement lists;
\(^\ddagger\)Qwen2.5's two sites were taken from the screen's descriptive
audit and fixed before its factorial sample, and the depth sweep covers
every layer.}
\label{tab:model-panel}
\end{table}

\paragraph{Panel.}  Each model reads the unchanged prompts with its own
tokenizer.  Models trained with a beginning-of-sequence token are run with
it (marked in Table~\ref{tab:model-panel}), since that token serves as an
attention sink \citep{xiao2024efficient}.

\paragraph{Binding task.}  Keys are first names; key-valued facts read
``Alice's friend is Bob.'', number-valued facts ``Bob lives in apartment
39.'', and the query follows the friends to an apartment number.  Values
stay two-digit, so the candidate readout is unchanged.

\paragraph{Pools and construction.}
The \emph{crossed assay} evaluates each crossing on graphs never used for
selection.  The \emph{common assay} draws disjoint training, validation, and
test pools (Table~\ref{tab:evidence-chain}).  Within each identity a
strength-two orthogonal array crosses the donor's key \(c\), answer \(a\),
and row \(p=(c+a)\bmod 3\), so every pair of factors is crossed once; the
cells whose key and row differ enter the factorial, and a second arm keeps
one donor state fixed across three permutations of the recipient's values.
All candidates share a tens token, so we feed the shared prefix and read the
unit-token logits.  Absolute selectivities differ between the two assays
(Table~\ref{tab:common-scale}), so comparisons stay within an assay.

\begin{table}[htbp]
\centering
\footnotesize
\setlength{\tabcolsep}{4pt}
\begin{tabular}{lrrr}
\toprule
Quantity (logits) & Crossed assay & Common assay & Normalized \\
\midrule
Recipient margin, \(A_r\) vs.\ runner-up & 1.45 & 1.71 \(\pm\) 0.21 & 1 \\
Mapping selectivity, \(F\) & 2.37 & 1.48 \(\pm\) 0.29 & 0.88 \(\pm\) 0.21 \\
Key selectivity, \(F\) & 1.02 & 1.14 \(\pm\) 0.10 & 0.66 \(\pm\) 0.09 \\
Mapping selectivity, pre-L39 & --- & 1.12 \(\pm\) 0.20 & 0.67 \(\pm\) 0.15 \\
Key selectivity, pre-L39 & --- & 0.85 \(\pm\) 0.08 & 0.50 \(\pm\) 0.07 \\
\(F\) minus pre-L39, mapping & --- & 0.36 \(\pm\) 0.09 & --- \\
\(F\) minus pre-L39, key & --- & 0.28 \(\pm\) 0.03 & --- \\
\bottomrule
\end{tabular}
\caption{Scale across assays.  Normalized values divide selectivity by the
recipient's own \(A_r\)-versus-runner-up margin at the identity level.  The
pre-L39 whole-residual transplant is the matched-layer control.}
\label{tab:common-scale}
\end{table}

\FloatBarrier
\section{Crossed assay validation}
\label{app:crossed}

\begin{table}[htbp]
 \centering\small
 \begin{tabular}{lccc}
 \toprule
 & \multicolumn{3}{c}{Donor's decisive row at position}\\
 \cmidrule(lr){2-4}
 Donor's key & 1 (Alice's fact) & 2 (Bob's fact) & 3 (Carol's fact)\\
 \midrule
 Alice & \textcolor{black!45}{Garden / Garden} & Garden / Kitchen & Garden / Hall\\
 Bob   & Kitchen / Garden & \textcolor{black!45}{Kitchen / Kitchen} & Kitchen / Hall\\
 Carol & Hall / Garden & Hall / Kitchen & \textcolor{black!45}{Hall / Hall}\\
 \bottomrule
 \end{tabular}
 \caption{\textbf{The factorial separates the key from the position.}
 The recipient binds Alice, Bob, and Carol to garden, kitchen, and hall,
 whose facts occupy rows 1, 2, and 3.  Each cell lists the location key-based
 retrieval predicts / the location row-based retrieval predicts.  On the
 diagonal (gray) the two coincide; on the six off-diagonal cells they differ,
 and the remaining third location is the reference candidate.}
 \label{tab:factorial}
\end{table}

\begin{figure}[t]
  \centering
  \resizebox{0.76\linewidth}{!}{
\begin{tikzpicture}[
  font=\scriptsize,
  row/.style={draw=black!45, rectangle, minimum width=1.55cm, minimum height=0.30cm,
              inner sep=1.5pt, outer sep=0pt, fill=white, font=\scriptsize},
  filler/.style={row, text=black!45},
  hlq/.style={row, fill=orange!30},
  hlA/.style={row, fill=orange!30},
  hlB/.style={row, fill=teal!25},
  hlS/.style={row, fill=red!22},
  hlC/.style={row, fill=blue!10},
  title/.style={font=\scriptsize\bfseries, anchor=west},
  lab/.style={font=\scriptsize, anchor=south, inner sep=1pt},
  pred/.style={draw=black!60, rounded corners=2pt, inner sep=3pt, align=left,
               font=\scriptsize, text width=4.2cm, fill=black!3},
  arr/.style={-{Latex[length=1.6mm]}, thick},
  every node/.style={outer sep=0pt},
]

\begin{scope}
  \node[title] at (0,0.35) {(a) Fixed-donor crossing};
  \node[lab] at (0.85,-0.05) {Donor};
  \node[hlq] (a-d1) at (0.85,-0.30) {$q \to m$};
  \node[hlA] (a-d2) at (0.85,-0.60) {$m \to A$};
  \node[filler] (a-d3) at (0.85,-0.90) {$r \to n$};
  \node[filler] (a-d4) at (0.85,-1.20) {$n \to C$};
  \node[filler] (a-d5) at (0.85,-1.50) {$\cdots$};
  \node[font=\scriptsize\itshape] at (0.85,-1.85) {query $q$};
  \node[lab] at (3.35,-0.05) {Recipients $j=1,2,3$};
  \node[hlB] (a-r1) at (3.35,-0.30) {$m \to B_1$};
  \node[hlC] (a-r1b) at (3.35,-0.60) {$r \to n \to C$};
  \node[hlB] (a-r2) at (3.35,-1.05) {$m \to B_2$};
  \node[hlC] (a-r2b) at (3.35,-1.35) {$r \to n \to C$};
  \node[hlB] (a-r3) at (3.35,-1.80) {$m \to B_3$};
  \node[hlC] (a-r3b) at (3.35,-2.10) {$r \to n \to C$};
  \node[font=\scriptsize\itshape] at (3.35,-2.45) {query $r$};
  \coordinate (a-src) at (1.70,-0.60);
  \draw[arr] (a-src) -- (a-r1.west |- a-r1.south);
  \draw[arr] (a-src) -- (a-r2.west |- a-r2.south);
  \draw[arr] (a-src) -- (a-r3.west |- a-r3.south);
  \node[font=\tiny, align=center, anchor=north] at (1.95,-1.95) {one donor\\state};
\end{scope}

\begin{scope}[shift={(4.75,0)}]
  \node[title] at (0,0.35) {(b) Fixed-recipient crossing};
  \node[lab] at (0.85,-0.05) {Donor $j{=}1$};
  \node[hlq] (b-d1) at (0.85,-0.30) {$q_1 \to m_1$};
  \node[filler] (b-d2) at (0.85,-0.60) {$\cdots$};
  \node[hlA] (b-d3) at (0.85,-0.90) {$m_1 \to A$};
  \node[filler] (b-d4) at (0.85,-1.20) {$\cdots$};
  \node[filler] (b-d5) at (0.85,-1.50) {$\cdots$};
  \node[font=\scriptsize\itshape] at (0.85,-1.85) {query $q_1$};
  \node[lab] at (3.35,-0.05) {Recipient (fixed)};
  \node[hlB] (b-r1) at (3.35,-0.30) {$m_1 \to B_1$};
  \node[filler] (b-r2) at (3.35,-0.60) {$q_1 \to \text{sink}$};
  \node[hlS] (b-r3) at (3.35,-0.90) {$m_2 \to B_2$};
  \node[row] (b-r4) at (3.35,-1.20) {$m_3 \to B_3$};
  \node[hlC] (b-r5) at (3.35,-1.50) {$r \to n \to C$};
  \node[font=\scriptsize\itshape] at (3.35,-1.85) {query $r$};
  \draw[arr] (1.70,-0.30) -- (2.50,-0.30);
  \node[font=\tiny, anchor=north] at (2.10,-0.36) {state};
  \draw[densely dashed, black!60] (b-d3.east) -- (b-r3.west);
  \node[font=\tiny, anchor=north] at (2.10,-0.96) {same row};
  \node[font=\tiny, align=center, text width=4.1cm, anchor=north] at (2.10,-2.15)
    {Donors $j{=}2,3$ analogous: row $m_j{\to}A$ sits where the recipient has
     $m_{j+1}{\to}B_{j+1}$.};
\end{scope}

\begin{scope}[shift={(0,-3.85)}]
  \node[title] at (0,0.35) {(c) Key $\times$ position factorial};
  \def\cs{0.60} 
  \def\cx{1.35} \def\gy{-0.75}
  \node[font=\scriptsize, anchor=south] at (\cx+1.5*\cs, \gy+0.26) {donor row position};
  \node[font=\scriptsize, rotate=90, anchor=south] at (\cx-0.62, \gy-1.5*\cs) {donor content};
  \foreach \j/\jl in {0/1,1/2,2/3} {
    \node[font=\tiny, anchor=south] at (\cx+\j*\cs+0.5*\cs, \gy+0.02) {\jl};
    \node[font=\tiny, anchor=east] at (\cx-0.06, \gy-\j*\cs-0.5*\cs) {$m_{\jl}$};
  }
  \foreach \i in {0,1,2} {
    \foreach \j in {0,1,2} {
      \pgfmathtruncatemacro{\ii}{\i+1}
      \pgfmathtruncatemacro{\jj}{\j+1}
      \ifnum\i=\j
        \node[draw=black!45, fill=black!12, minimum size=\cs cm, inner sep=0pt,
              font=\tiny, text=black!55, anchor=north west]
          at (\cx+\j*\cs, \gy-\i*\cs) {agree};
      \else
        \node[draw=black!45, minimum size=\cs cm, inner sep=0pt, font=\tiny,
              align=center, anchor=north west]
          at (\cx+\j*\cs, \gy-\i*\cs)
          {\textcolor{teal!70!black}{$B_{\ii}$}\\\textcolor{red!70!black}{$B_{\jj}$}};
      \fi
    }
  }
  \node[font=\tiny, anchor=north, align=center, text width=4.1cm] at (2.10,-2.60)
    {Grey diagonal cells, where both accounts agree, are excluded.};
\end{scope}

\begin{scope}[shift={(4.75,-3.85)}]
  \node[title] at (0,0.35) {Predicted answer under each account};
  \node[pred, anchor=north west] at (0,0.05) {%
    \textbf{(a)}\; fixed answer: $A$ (same for all $j$);\;
    position: value in the aligned row;\;
    content: $B_j$ (follows the recipient).};
  \node[pred, anchor=north west] at (0,-1.35) {%
    \textbf{(b)}\; fixed answer: $A$;\;
    position: $B_{j+1}$ (the aligned row);\;
    content: $B_j$ (the donor's intermediate key).};
  \node[pred, anchor=north west] at (0,-2.65) {%
    \textbf{(c)}\; content: \textcolor{teal!70!black}{$B_i$} (donor's $m_i$; row);\;
    position: \textcolor{red!70!black}{$B_j$} (occupied row; column);\;
    third value $B_k$: neither.};
\end{scope}

\end{tikzpicture}}
  \caption{\textbf{Three crossed constructions.}  (a) One donor state enters
  three recipients that bind the same key \(m\) to \(B_1,B_2,B_3\).  (b)
  Three donors whose queries reach different keys \(m_j\) enter one recipient;
  the donor row \(m_j\to A\) is placed where the recipient holds
  \(m_{j+1}\to B_{j+1}\), and the recipient maps the donor's query \(q_j\)
  away from \(B_j\) (to a dead end in most cells, Appendix~\ref{app:common}),
  so neither the query token nor the row position predicts \(B_j\).  (c) Donor key and donor row position vary independently;
  diagonal cells are excluded.  Colors: orange donor path, teal recipient
  bindings under test, red adversarially aligned row, blue the recipient's
  own path; the lower-right panel lists each account's predicted answer.}
  \label{fig:method}
\end{figure}
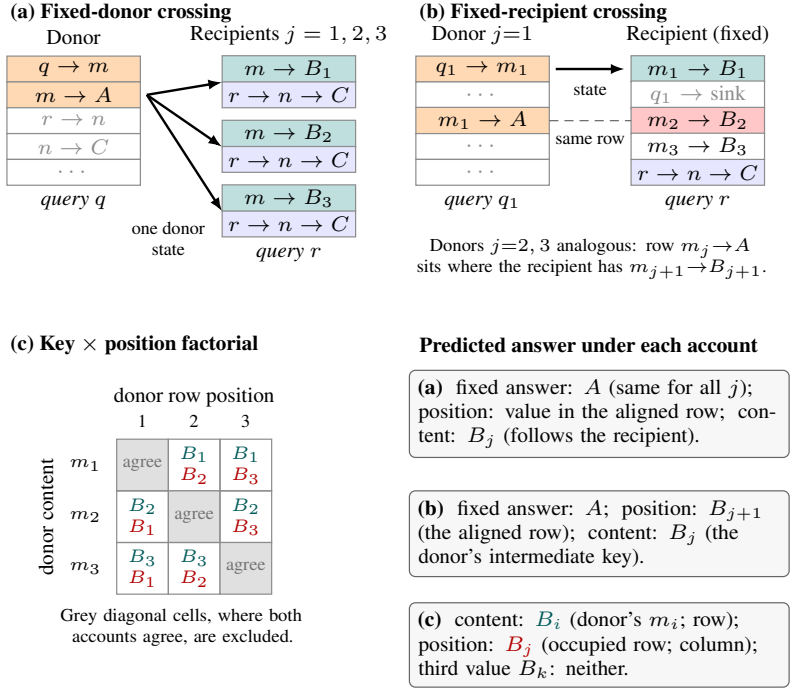

\paragraph{Complete construction example.}
In the literal prompts behind Figure~\ref{fig:handoff}b, the donor's query
reaches the key \texttt{CFG}, which the recipient still binds to a different
value, while the donor's query token leads nowhere in the recipient, so the
query token alone cannot predict the key's value.  Following the donor's query through the recipient's own facts reaches the
key's candidate in 57 of the pool's 1,800 cells, and excluding those leaves
the key effect unchanged: 1.13 against 1.10 at Nemotron-H's pre-L40 site,
and 1.07 against 1.07 at Granite-H's pre-L35.  Serialized prompts for every construction are in the
supplement.

\begin{table}[!htb]
\centering
\scriptsize
\setlength{\tabcolsep}{3.5pt}
\begin{tabular}{llrr}
\toprule
 & & Nemotron-H, pre-L40 & Granite-H, pre-L35 \\
\midrule
\multirow{5}{*}{\shortstack[l]{Fixed\\donor}}
 & Mapping selectivity (logits) & 2.37 \(\pm\) 0.21 & 2.48 \(\pm\) 0.22 \\
 & Mapped vs.\ donor answer \(A\) (logits) & 1.57 \(\pm\) 0.19 & 1.67 \(\pm\) 0.24 \\
 & Mapped vs.\ recipient answer \(A_r\) (logits) & 4.17 \(\pm\) 0.27 & 4.84 \(\pm\) 0.31 \\
 & Tracking rate (\%) & 80.8 \(\pm\) 4.5 & 86.0 \(\pm\) 3.8 \\
 & Top-1 rate among seven (\%) & 85.8 \(\pm\) 4.5 & 70.8 \(\pm\) 5.5 \\
\midrule
\multirow{5}{*}{\shortstack[l]{Fixed\\recipient}}
 & Key selectivity (logits) & 1.02 \(\pm\) 0.12 & 0.99 \(\pm\) 0.14 \\
 & Key vs.\ slot answer \(B_{j+1}\) (logits) & 0.86 \(\pm\) 0.14 & 0.89 \(\pm\) 0.17 \\
 & Tracking rate (\%) & 52.8 \(\pm\) 3.6 & 52.8 \(\pm\) 3.5 \\
 & Top-1 rate among seven (\%) & 39.3 \(\pm\) 3.9 & 35.2 \(\pm\) 3.8 \\
 & Explicit-key reference, top-1 (\%)\(^\ast\) & 98.8 & 99.3 \\
\bottomrule
\end{tabular}
\caption{Crossed assay: 200 held-out graphs (600 crosses) per construction
and model, with 95\% intervals; chance tracking is 33.3\%.  The unmodified
Nemotron-H recipient's margin between \(A_r\) and its runner-up is 1.45
logits.  \(^\ast\)Accuracy when the recipient is queried from \(m_j\)
directly.  Unmodified accuracies are in Appendix Table~\ref{tab:baselines}.}
\label{tab:crossed}
\end{table}

\begin{table}[!htb]
\centering
\small
\setlength{\tabcolsep}{4pt}
\begin{tabular}{llrr}
\toprule
 & & Nemotron-H & Granite-H \\
\midrule
\multirow{4}{*}{\shortstack[l]{Fixed-donor\\crossing}}
 & Unmodified donor accuracy (\%) & 88.5 & 82.0 \\
 & Unmodified recipient accuracy (\%) & 71.7 & 84.7 \\
 & Donor prefers \(A\) over \(A_r\) (\%) & 95.5 & --- \\
 & Recipient prefers \(A_r\) over \(A\) (\%) & 92.2 & --- \\
\midrule
\multirow{4}{*}{\shortstack[l]{Fixed-recipient\\crossing}}
 & Unmodified donor accuracy (\%) & 85.5 & 82.0 \\
 & Unmodified recipient accuracy (\%) & 82.0 & 84.7 \\
 & Donor prefers \(A\) over \(A_r\) (\%) & 93.7 & --- \\
 & Recipient prefers \(A_r\) over \(A\) (\%) & 95.5 & --- \\
\midrule
\multirow{3}{*}{\shortstack[l]{Granite-H\\L35 attention restored}}
 & Mapping selectivity, total / remaining & --- & 2.58 / \(-0.05\) \\
 & Key selectivity, total / remaining & --- & 0.87 / \(-0.11\) \\
 & Key vs.\ slot, total / remaining & --- & 0.70 / \(-0.12\) \\
\bottomrule
\end{tabular}
\caption{Unmodified accuracies (seven-candidate top-1 on the held-out graphs)
and Granite-H layer-level mediation for the crossed assay.  Granite-H used one
held-out sample for both constructions; its mediation totals come from the
mediation sample and differ slightly from Table~\ref{tab:crossed}.}
\label{tab:baselines}
\end{table}

\section{Depth profiles and the resolving layer}
\label{app:profile}
\label{app:panel}

Figure~\ref{fig:profile} shows the main depth profiles.  The tables below
give their contrasts and classifications at individual layer inputs,
followed by the second-format and probe comparisons.

\begin{table}[htbp]
\centering\small
\begin{tabular}{lrrr}
\toprule
Nemotron-H & Attention & Recurrent, before L40 & Recurrent, after L40 \\
\midrule
Arrow task & 0.947 & 0.036 & 0.016 \(\pm\) 0.005 \\
Word task & 0.862 & 0.046 & 0.092 \(\pm\) 0.010 \\
Numeric twin & 0.936 & 0.036 & 0.028 \(\pm\) 0.005 \\
\bottomrule
\end{tabular}
\caption{Where the rise in the donor-answer effect falls, by task.  Shares
of the whole rise, with 95\% identity-bootstrap intervals on the
post-attention share, the one that separates the tasks: it is higher on the
word task than on the arrow task by 0.075 \(\pm\) 0.011 and than on the
numeric twin by 0.064 \(\pm\) 0.012.  The word task is the one on which
Section~\ref{sec:suffix} finds the memory read and the numeric twin the one
on which it does not.  Within the word task's post-attention band the rise
concentrates in L44 (0.044 of the whole, against 0.024, 0.017, 0.006 and
0.001 for L42, L46, L50 and L48), the layer whose output clamp costs the
post-attention lookup most among all five clamped individually, on two
identity pools (Appendix~\ref{app:suffix-retrieval}).}
\label{tab:word-share}
\end{table}

\paragraph{Identities the model answers correctly.}
Every score in the paper averages over all 200 identities of its pool,
whether or not the unmodified model answers them.  Splitting Nemotron-H's
pools by that leaves none of the effects smaller on the identities it
answers.  For the key effect at the pre-L40 input, where the donor and the
recipient prompts must both be right and each is on 84\% of identities, the
42 that qualify give 1.34 \(\pm\) 0.22 and the other 158 give 1.07 \(\pm\) 0.12, against 1.13 over all of them.  For the memory read, where the test
prompt and both memory prompts must be right and the three are right on
69\% of prompts, the 35 that qualify give 1.08 \(\pm\) 0.13 and 0.82 \(\pm\) 0.11 and the other 165 give 0.96 \(\pm\) 0.06 and 0.68 \(\pm\) 0.05,
against 0.98 and 0.70 over all.  Nano, whose read is the smallest of the four, splits
the same way in \(\sigma_{\rm test}\) on the word task, where 98 of its 200
identities qualify (0.30 \(\pm\) 0.05 against 0.24 \(\pm\) 0.04 for the
rest) and on the paragraph version, where 30 do (0.17 \(\pm\) 0.04 against
0.12 \(\pm\) 0.02).  The effect is therefore present in both halves of both models, and
the averages we report are the conservative ones; the subgroups are nested
in the full pool, so these are descriptive splits rather than a test.

\paragraph{How much the alignment could have failed.}
The resolving layer and the peak of the key effect are read from the same
profiles, so the alignment would be empty if the rise occupied a single
layer.  It does not: the span from a tenth to nine tenths of the cumulative
rise covers three to twelve layers (median seven,
Table~\ref{tab:rise-window}), with two to nine attention layers inside it.
Profiles of the same model on the two tasks share their rise spans, and
three pairs of checkpoints are siblings, so we count the nine model
families rather than the 23 profiles.  A position-wise MLP or MoE layer
maps the transplanted residual deterministically, so the transplants on
either side of one are identical and its increment is exactly zero; such a
layer cannot carry the largest step, and the null is taken over the layers
that can.  Drawing the largest step uniformly from the token-mixing layers
in a family's span would place it at the key peak in 1.7 of the nine;
drawing it from the attention layers in the span, the more generous null,
gives 3.1.  All nine coincide.  What the alignment establishes is where the
usable key is: in every profile we measured it is largest at the input of
the layer that spends it, rather than at some earlier layer where the key
is already available.  Which components do the spending is what the
interventions of Section~\ref{sec:mechanism} address.

\paragraph{Depth profile.}
We repeat \(F\) at the answer-position residual entering every layer of
each model in Table~\ref{tab:panel}, on the common-assay test identities,
and classify each layer as in Section~\ref{sec:setup}.  The profile is
descriptive: per-layer intervals carry no multiplicity adjustment, and no
layer is selected by outcome.

\begin{table}[htbp]
\centering
\scriptsize
\setlength{\tabcolsep}{3.5pt}
\resizebox{\textwidth}{!}{%
\begin{tabular}{lrrrl}
\toprule
Site & Key \(\kappa\) & Position \(\pi\) & \(\kappa-\pi\) & Class \\
\midrule
Nemotron-H pre-L40 & 1.13 \(\pm\) 0.12 & 0.36 \(\pm\) 0.10 & 0.77 \(\pm\) 0.11 & Key-dominant \\
Nemotron-H pre-L39 & 0.84 \(\pm\) 0.09 & 0.28 \(\pm\) 0.07 & 0.56 \(\pm\) 0.09 & Key-dominant \\
Qwen2.5 pre-L22 & 0.86 \(\pm\) 0.16 & 0.87 \(\pm\) 0.14 & \(-\)0.02 \(\pm\) 0.14 & Balanced \\
Qwen2.5 pre-L26 & \(-\)0.63 \(\pm\) 0.12 & 0.02 \(\pm\) 0.11 & \(-\)0.65 \(\pm\) 0.09 & Neither\(^\dagger\) \\
Granite-H pre-L35 & 1.07 \(\pm\) 0.13 & 0.25 \(\pm\) 0.09 & 0.82 \(\pm\) 0.10 & Key-dominant \\
Nemotron-H pre-L40 (colon format) & 0.81 \(\pm\) 0.12 & 0.31 \(\pm\) 0.08 & 0.49 \(\pm\) 0.09 & Key-dominant \\
\bottomrule
\end{tabular}}
\caption{Key \(\times\) position factorial, in logits, on 200 held-out
graphs (1,200 off-diagonal cells) per site.  \(\kappa\) and \(\pi\) are the
key-vs-reference and position-vs-reference contrasts of Eq.~\ref{eq:contrasts}; they carry 97.5\%
intervals; the difference carries the 90\% interval used for the
prespecified \(\pm0.30\) balanced region.  Every key-dominant row is
also dual-supported.  The Granite-H class and the second-format class were
predicted in advance.  \(^\dagger\)The key is suppressed below the reference
candidate and position does not reliably exceed it.}
\label{tab:factorial-sites}
\end{table}

\begin{table}[htbp]
\centering
\scriptsize
\setlength{\tabcolsep}{3pt}
\begin{tabular}{llll}
\toprule
Model & No class & Balanced & Key-dominant \\
\midrule
Nemotron-H & pre-L0--L26; pre-L41--L51 (fixed answer) & pre-L27, L28, L31--L33 & pre-L29, L30, L34--L40 \\
Granite-H & pre-L0--L24; pre-L36, L39 (fixed answer) & --- & pre-L25\(^\S\), L26--L35 \\
Qwen2.5 & pre-L0--L17; pre-L24, L26, L27 (fixed answer) & pre-L18, L19, L21--L23 & pre-L20 \\
\bottomrule
\end{tabular}
\caption{Prespecified class at every layer input.  Position-dominant labels
occur only in the fixed-answer regime (Granite-H pre-L37, L38; Qwen2.5
pre-L25), where the key lies below the reference candidate and the positional
contrast is about 0.1 logits.  \(^\S\)Key-dominant without dual
support.  Source and recipient seven-way accuracies were 84/84\%
(Nemotron-H), 77/81\% (Granite-H), and 89/91\% (Qwen2.5).  Consecutive
layers separated only by a position-wise MLP return identical values because
the MLP is a deterministic function of the transplanted final-token state.}
\label{tab:profile-bands}
\end{table}

\paragraph{Second format.}  The factorial is regenerated under a second
surface form (\texttt{XYZ: ABC} facts under a \texttt{Mappings:} header, a
different question, and the prefix \texttt{Result:}); the prediction that
pre-L40 keeps its class was fixed before the run and holds
(Table~\ref{tab:factorial-sites}).

\begin{figure}[htbp]
  \centering
  \includegraphics[width=0.55\linewidth]{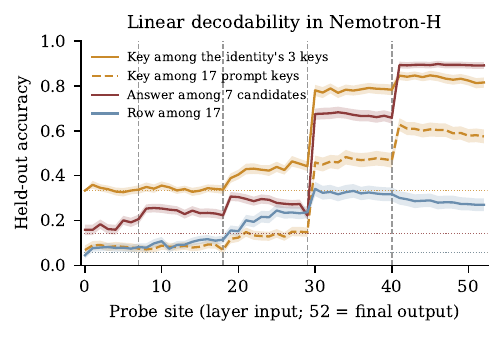}
  \caption{Ridge-probe decodability from the donor's answer-position
  residual at each layer input in Nemotron-H (site 52 is the final layer
  output), with 95\% identity-clustered intervals; dotted lines mark chance
  and dashed lines the attention layers.  Shuffled-label probes stay at
  chance at every site.}
  \label{fig:probes}
\end{figure}

\paragraph{The key is decodable long before it is read.}
Ridge probes fitted on the training identities and selected on the
validation identities retrieve the intermediate key from the donor's
residual well above chance ten layers before the causal effect peaks, and
more often than the transplant tracks it (Figure~\ref{fig:probes});
shuffled-label probes stay at chance, and decodability bounds a linear
reader only.  The key also survives the fall the resolving layer produces:
at the layer input after it the probe recovers the key in 0.62 to 0.86 of
identities across the twelve models, against a chance level of one in
three.

\begin{table}[htbp]
\centering
\scriptsize
\setlength{\tabcolsep}{2.5pt}
\renewcommand{\arraystretch}{0.92}
\resizebox{\textwidth}{!}{%
\begin{tabular}{llrlrrrr}
\toprule
Model & Type & Layers & Key peak (class) & \(C\), arrow & \(C\), binding & \(e^{H}\), arrow / binding & Attn.\ share \\
\midrule
Nemotron-H-8B & Sequential hybrid & 52 (4 attn.) & pre-L40 (key) & 0.68 \(\pm\) 0.03 & 0.69 \(\pm\) 0.02 & 1.9 / 1.9 & 0.95 \\
Nemotron-H-4B & Sequential hybrid & 52 (4 attn.) & pre-L40 (key) & 0.63 \(\pm\) 0.03 & 0.68 \(\pm\) 0.02 & 1.9 / 1.8 & 0.94 \\
Granite-H & Sequential hybrid & 40 (4 attn.) & pre-L35 (key) & 0.70 \(\pm\) 0.02 & 0.55 \(\pm\) 0.02 & 1.8 / 2.0 & 0.96 \\
\midrule
Falcon-H1-3B & Parallel hybrid & 32 (all) & pre-L21 (key) & 0.59 \(\pm\) 0.02 & 0.56 \(\pm\) 0.02 & 4.6 / 4.8 & 1 \\
Falcon-H1-7B & Parallel hybrid & 44 (all) & pre-L29 (balanced) & 0.70 \(\pm\) 0.02 & 0.73 \(\pm\) 0.01 & 3.2 / 3.0 & 1 \\
\midrule
Qwen2.5-7B & Dense & 28 & pre-L22 (balanced) & 0.61 \(\pm\) 0.02 & 0.54 \(\pm\) 0.02 & 2.9 / 3.2 & 1 \\
Qwen2.5-14B & Dense & 48 & pre-L36 (position) & 0.38 \(\pm\) 0.02 & 0.36 \(\pm\) 0.01 & 5.5 / 5.9 & 1 \\
Mistral-7B & Dense & 32 & pre-L19 (key) & 0.44 \(\pm\) 0.02 & 0.43 \(\pm\) 0.02 & 3.4 / 4.0 & 1 \\
Llama-3.1-8B & Dense & 32 & pre-L16 (key) & 0.35 \(\pm\) 0.01 & 0.32 \(\pm\) 0.01 & 7.1 / 8.0 & 1 \\
Gemma-2-9B & Dense & 42 & pre-L28 (key) & 0.65 \(\pm\) 0.02 & 0.68 \(\pm\) 0.01 & 4.1 / 3.6 & 1 \\
\midrule
RecurrentGemma-9B & Sequential hybrid & 38 (12 attn.) & pre-L29 (key) & 0.76 \(\pm\) 0.02 & Gate fail & 2.0 / --- & 0.96 \\
Nemotron-3-Nano-30B & Sequential hybrid & 52 (6 attn.) & pre-L33 (key) & 0.69 \(\pm\) 0.02 & 0.69 \(\pm\) 0.02 & 1.7 / 1.9 & 0.88 \\
\bottomrule
\end{tabular}}
\caption{The model panel behind Figure~\ref{fig:concentration}: concentration \(C\), the largest single layer's share of the rise in the
donor-answer effect over all layers (95\% intervals), on both tasks; the
effective attention count \(e^{H}\) carrying that rise;
attention's share of the rise (the summed positive increments at attention
layers divided by the sum over all layers); and the site and class of the
largest key effect \(\kappa\) on the arrow task.}
\label{tab:panel}
\end{table}

\begin{figure}[!htb]
  \centering
  \includegraphics[width=\linewidth]{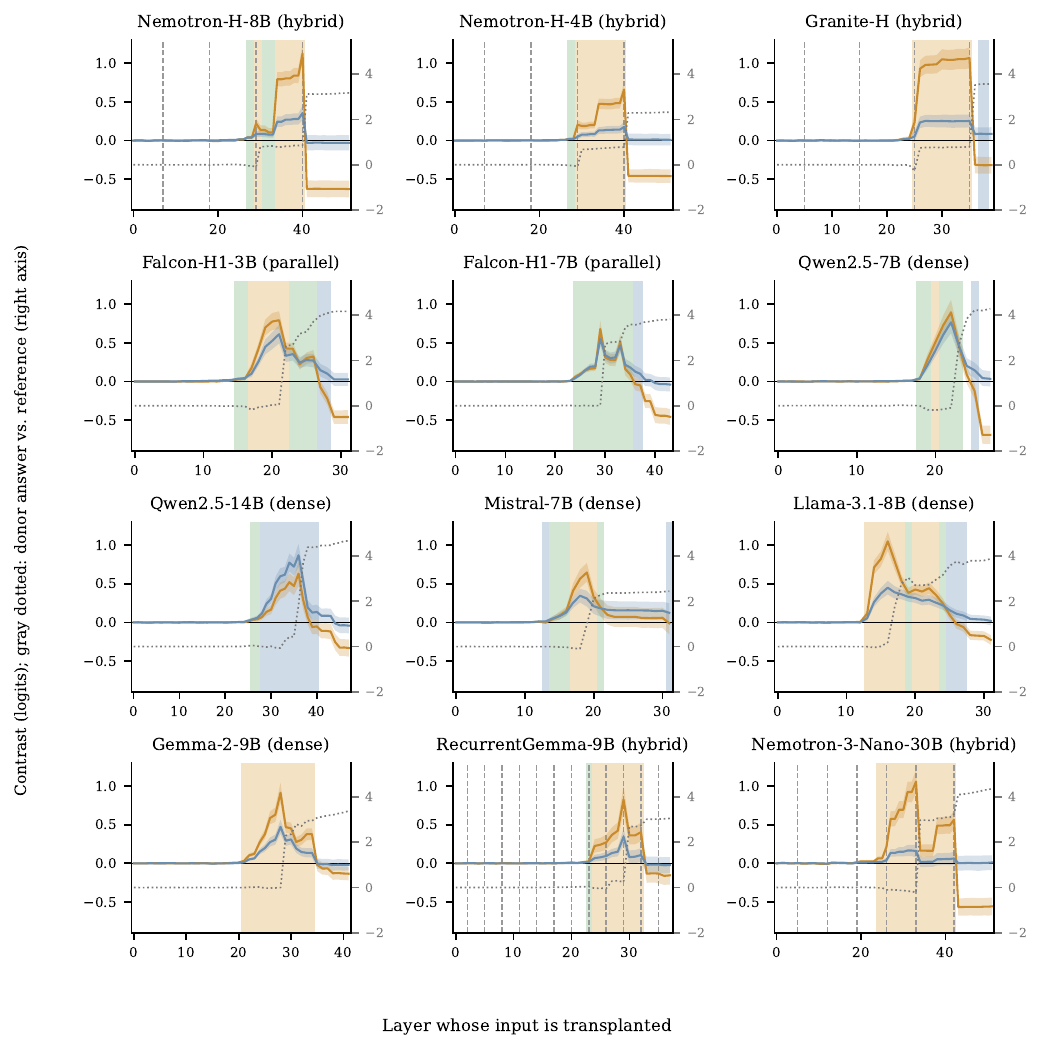}
  \caption{Depth profiles of the twelve models in Table~\ref{tab:panel}, in
  the format of Figure~\ref{fig:profile}.  Dashed lines mark attention
  layers in the sequential hybrids; the parallel hybrids and dense models
  have attention in every layer.}
  \label{fig:panel}
\end{figure}

\begin{figure}[!htb]
  \centering
  \includegraphics[width=\linewidth]{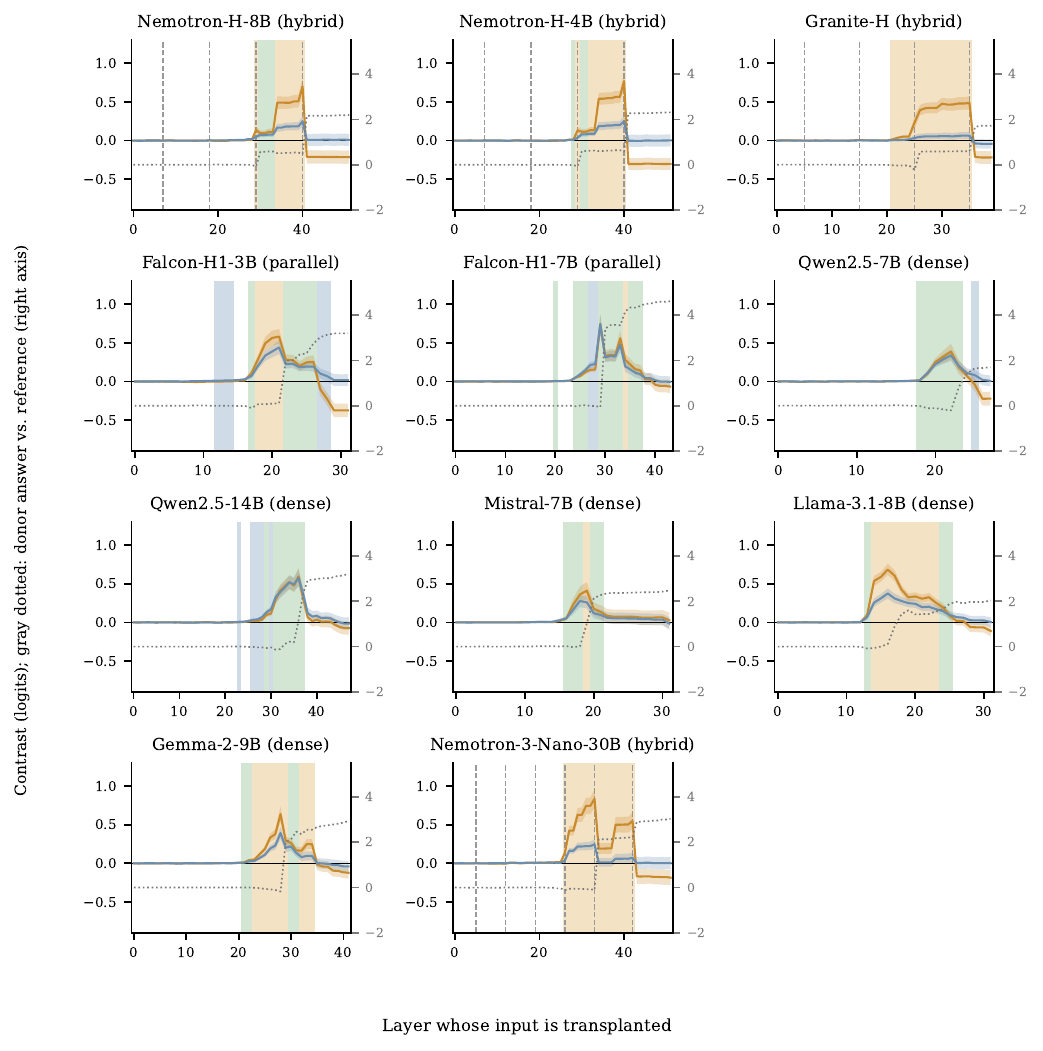}
  \caption{Depth profiles on the binding task of
  Section~\ref{sec:setup} for the models of Table~\ref{tab:panel}
  that passed the binding-task gate, in the format of
  Figure~\ref{fig:panel}.}
  \label{fig:panel-binding}
\end{figure}
\FloatBarrier   

\paragraph{Peak alignment.}
In all 23 eligible profiles, the input to the layer with the largest
donor-answer increment carries the peak key effect.  The relationship was
documented on the first 21 profiles before Nano was evaluated, and recurs on
both Nano profiles.  Resampling the identities of each profile
(Table~\ref{tab:peak-bootstrap}), the resolving layer never changes and the
peak falls at its input in nearly every draw, under both contrasts.  When
two layer inputs carry the same key effect because only a position-wise MLP
separates them, the earlier one is taken as the peak.

\begin{table}[htbp]
\centering
\scriptsize
\setlength{\tabcolsep}{4pt}
\begin{tabular}{llccc}
\toprule
Task & Model & Peak = resolver & Coincide (\%) & Peak margin over runner-up \\
\midrule
Arrow & Nemotron-H-8B & L40 & 100.0 & 0.29 \(\pm\) 0.04 (L38) \\
Arrow & Nemotron-H-4B & L40 & 100.0 & 0.18 \(\pm\) 0.02 (L39) \\
Arrow & Granite-H & L35 & 99.4 & 0.01 \(\pm\) 0.01 (L33) \\
Arrow & RecurrentGemma-9B & L29 & 100.0 & 0.40 \(\pm\) 0.07 (L28) \\
Arrow & Nemotron-3-Nano-30B & L33 & 100.0 & 0.14 \(\pm\) 0.04 (L31) \\
Arrow & Falcon-H1-3B & L21 & 97.8 & 0.02 \(\pm\) 0.01 (L20) \\
Arrow & Falcon-H1-7B & L29 & 100.0 & 0.17 \(\pm\) 0.05 (L33) \\
Arrow & Qwen2.5-7B & L22 & 100.0 & 0.17 \(\pm\) 0.03 (L21) \\
Arrow & Qwen2.5-14B & L36 & 100.0 & 0.11 \(\pm\) 0.06 (L34) \\
Arrow & Mistral-7B & L19 & 88.3 & 0.08 \(\pm\) 0.07 (L18) \\
Arrow & Llama-3.1-8B & L16 & 100.0 & 0.23 \(\pm\) 0.04 (L15) \\
Arrow & Gemma-2-9B & L28 & 100.0 & 0.28 \(\pm\) 0.04 (L27) \\
\midrule
Binding & Nemotron-H-8B & L40 & 100.0 & 0.19 \(\pm\) 0.03 (L39) \\
Binding & Nemotron-H-4B & L40 & 100.0 & 0.20 \(\pm\) 0.03 (L38) \\
Binding & Granite-H & L35 & 97.0 & 0.01 \(\pm\) 0.01 (L33) \\
Binding & Nemotron-3-Nano-30B & L33 & 100.0 & 0.09 \(\pm\) 0.02 (L32) \\
Binding & Falcon-H1-3B & L21 & 99.4 & 0.02 \(\pm\) 0.01 (L20) \\
Binding & Falcon-H1-7B & L29 & 100.0 & 0.18 \(\pm\) 0.06 (L33) \\
Binding & Qwen2.5-7B & L22 & 100.0 & 0.07 \(\pm\) 0.02 (L21) \\
Binding & Qwen2.5-14B & L36 & 100.0 & 0.08 \(\pm\) 0.04 (L34) \\
Binding & Mistral-7B & L19 & 97.9 & 0.05 \(\pm\) 0.04 (L18) \\
Binding & Llama-3.1-8B & L16 & 100.0 & 0.08 \(\pm\) 0.02 (L17) \\
Binding & Gemma-2-9B & L28 & 100.0 & 0.26 \(\pm\) 0.04 (L27) \\
\bottomrule
\end{tabular}

\caption{Stability of peak alignment under resampling of identities (2,000
replicates per profile).  ``Coincide'' is the share of replicates in which
the layer input with the largest key effect is the input of the layer with
the largest donor-answer increment.  The margin is the peak key effect minus the
largest key effect at any other layer input (the runner-up, in parentheses),
with its 95\% interval.}
\label{tab:peak-bootstrap}
\end{table}

\paragraph{Attention layers in the conversion band.}
Table~\ref{tab:rise-window} gives, for each profile, the span of layers
over which the donor-answer effect rises from 10\% to 90\% of its
cumulative positive increment, the attention layers inside that span, and
the effective attention count \(e^H\) of Eq.~\ref{eq:increment}.  The span
occupies a similar band of depth in every model, and \(e^H\) tracks the
attention layers inside it: the sequential hybrids have one or two there
and use them, the dense models have more and spread the rise across them.
RecurrentGemma-9B, with six attention layers in its second half, still
confines the rise to two of them.

\begin{table}[htbp]
\centering
\scriptsize
\setlength{\tabcolsep}{3pt}
\resizebox{\textwidth}{!}{%
\begin{tabular}{lrrlrrrrrlrrrrr}
\toprule
 & & & \multicolumn{6}{c}{Arrow task} & \multicolumn{6}{c}{Binding task} \\
\cmidrule(lr){4-9}\cmidrule(lr){10-15}
Model & Attn.\ blocks & Late attn. & Rise span & Depth & Attn.\ in span & \(e^{H}\) & \(\kappa\) in & Consumed & Rise span & Depth & Attn.\ in span & \(e^{H}\) & \(\kappa\) in & Consumed \\
\midrule
Nemotron-H-8B & 4 & 2 & L29--L40 & 0.57--0.78 & 2 & 1.9 & 1.13 & 1.55 & L29--L40 & 0.57--0.78 & 2 & 1.9 & 0.70 & 1.30 \\
Nemotron-H-4B & 4 & 2 & L29--L40 & 0.57--0.78 & 2 & 1.9 & 0.66 & 1.69 & L29--L40 & 0.57--0.78 & 2 & 1.8 & 0.77 & 1.39 \\
Granite-H & 4 & 2 & L25--L35 & 0.64--0.90 & 2 & 1.8 & 1.07 & 1.29 & L25--L35 & 0.64--0.90 & 2 & 2.0 & 0.49 & 1.43 \\
RecurrentGemma-9B & 12 & 6 & L26--L31 & 0.70--0.84 & 2 & 2.0 & 0.83 & 0.56 & --- & --- & --- & --- & --- & --- \\
Nemotron-3-Nano-30B & 6 & 3 & L33--L42 & 0.65--0.82 & 2 & 1.7 & 1.06 & 0.85 & L33--L42 & 0.65--0.82 & 2 & 1.9 & 0.83 & 0.77 \\
\midrule
Falcon-H1-3B & 31 & 15 & L21--L26 & 0.68--0.84 & 6 & 4.6 & 0.79 & 0.46 & L21--L26 & 0.68--0.84 & 6 & 4.8 & 0.58 & 0.52 \\
Falcon-H1-7B & 43 & 21 & L29--L33 & 0.67--0.77 & 5 & 3.2 & 0.68 & 0.54 & L29--L33 & 0.67--0.77 & 5 & 3.0 & 0.74 & 0.55 \\
\midrule
Qwen2.5-7B & 27 & 13 & L22--L24 & 0.81--0.89 & 3 & 2.9 & 0.90 & 0.49 & L22--L24 & 0.81--0.89 & 3 & 3.2 & 0.39 & 0.52 \\
Qwen2.5-14B & 47 & 23 & L33--L37 & 0.70--0.79 & 5 & 5.5 & 0.63 & 0.54 & L33--L37 & 0.70--0.79 & 5 & 5.9 & 0.60 & 0.47 \\
Mistral-7B & 31 & 15 & L18--L20 & 0.58--0.65 & 3 & 3.4 & 0.65 & 0.49 & L18--L20 & 0.58--0.65 & 3 & 4.0 & 0.41 & 0.57 \\
Llama-3.1-8B & 31 & 15 & L16--L24 & 0.52--0.77 & 9 & 7.1 & 1.05 & 0.24 & L16--L24 & 0.52--0.77 & 9 & 8.0 & 0.68 & 0.12 \\
Gemma-2-9B & 41 & 20 & L28--L34 & 0.68--0.83 & 7 & 4.1 & 0.91 & 0.49 & L28--L34 & 0.68--0.83 & 7 & 3.6 & 0.64 & 0.55 \\
\bottomrule
\end{tabular}
}
\caption{Conversion band per profile: the layer span carrying the middle
80\% of the cumulative positive rise in the donor-answer effect, its
position as a fraction of depth, the attention layers inside it, and the
effective attention count \(e^H\); ``\(\kappa\) in'' is the key effect
entering the resolving layer and ``Consumed'' its drop across that layer as
a fraction of that value.  ``Late attn.'' counts attention layers in the
second half of the network.  RecurrentGemma-9B's binding
run failed its gate.}
\label{tab:rise-window}
\end{table}

\paragraph{Repeated conversion in Nano.}
On both tasks Nano's key effect peaks at the input of attention layer L33,
which carries the largest share of the rise, with L42 carrying most of the
rest; recurrent layer L37 rebuilds the key between them
(Figure~\ref{fig:profile}).

\FloatBarrier
\section{Recurrent writers before attention}
\label{app:writers}

\paragraph{Writer swap.}
Table~\ref{tab:writer-swap} reports the writer-swap experiment on 200 fresh
identities disjoint from the common assay.  Each identity has one host donor
and guests with the host's row layout: key-swap guests matched on the
answer and the decisive row but reaching a different intermediate key, and
answer-swap guests matched on the key but with a different answer.  The host
is run with a guest's final-token output of a recurrent layer in place of
its own, the run continues to the resolving layer's input, and that state
is transplanted into the recipient; the same is done for matched
low-increment layers, for the nominated writers together, and for the
guest's whole residual after the writer.  Because a redirection fraction
cannot tell a removed host lookup from a created guest lookup, the table
reports the guest and host lookups separately, the shared-key lookup
retained under answer swaps, and the host's own logit shift.  Heads-only
and heads-clamped rows use the frozen head sets of
Section~\ref{sec:models} and Appendix~\ref{app:reader}, except in Nano,
whose head set at L33 was selected after this experiment ran, so its rows
use every head of that layer.  The residual row is an upper bound rather than a
rival account: it carries everything the run has computed up to that layer,
the nominated outputs among it, so it creates more of the guest's lookup
than those outputs alone.  What singles out the nominated recurrent layers
from the other recurrent layers is the matched comparison, where they leave
a tenth of the host's lookup and the matched layers leave seven tenths.  In every model a
recurrent-output swap creates part of the guest's lookup and same-key swaps
retain the lookup.  The frozen heads carry that effect in Nemotron-H, where
the heads-only rows reproduce the guest lookup of the full swap, and part
of it in Granite-H, where they carry 0.07 of the writer's 0.20.  The
heads-clamped rows are read from the guest and host columns rather than
from redirection: clamping removes the lookup instead of moving it, leaving
no guest lookup and almost no host lookup (0.00 and 0.08 in Nemotron-H),
while redirection, a fraction between the host and guest transplants, rises
because the run has left the host's end.  In Granite-H the two head
conditions do not divide the writer's effect between them: the clamped rows
keep more guest lookup than the heads-only rows in all three (0.14 against
0.07 for the single writer, 0.13 against 0.05 for the writers together, and
0.14 against 0.08 for the residual), so part of what the writer creates
reaches the answer without the frozen heads, and we do not have an account
of that path.

\begin{table}[htbp]
\centering
\scriptsize
\setlength{\tabcolsep}{3pt}
\resizebox{\textwidth}{!}{%
\begin{tabular}{lrrrrr}
\toprule
Condition & Redirection & Guest lookup & Host lookup & Shared key kept & Host frame \\
\midrule
\multicolumn{6}{l}{\textit{Nemotron-H-8B}: reader L40, 8 of 32 heads, gate 84/84\%} \\
Writer L33 swapped & 0.04 \(\pm\) 0.03 & 0.04 \(\pm\) 0.04 & 0.95 \(\pm\) 0.06 & 1.00 \(\pm\) 0.02 & \(-\)0.02 \(\pm\) 0.04 \\
Control L35 swapped & 0.13 \(\pm\) 0.02 & 0.14 \(\pm\) 0.02 & 0.87 \(\pm\) 0.05 & 1.00 \(\pm\) 0.01 & 0.13 \(\pm\) 0.04 \\
All writers L28/L33/L39 swapped & 0.58 \(\pm\) 0.05 & 0.31 \(\pm\) 0.09 & 0.10 \(\pm\) 0.20 & 0.99 \(\pm\) 0.03 & 0.32 \(\pm\) 0.14 \\
Control set L31/L35/L37 swapped & 0.26 \(\pm\) 0.03 & 0.26 \(\pm\) 0.03 & 0.72 \(\pm\) 0.09 & 1.00 \(\pm\) 0.02 & 0.24 \(\pm\) 0.05 \\
Residual after L33 replaced & 0.69 \(\pm\) 0.02 & 0.74 \(\pm\) 0.03 & 0.33 \(\pm\) 0.17 & 0.99 \(\pm\) 0.04 & 0.80 \(\pm\) 0.13 \\
\quad Writer swap, heads only & 0.04 \(\pm\) 0.05 & 0.04 \(\pm\) 0.03 & 0.95 \(\pm\) 0.10 & --- & --- \\
\quad Writer swap, heads clamped & 0.44 \(\pm\) 0.07 & 0.00 \(\pm\) 0.01 & 0.08 \(\pm\) 0.10 & --- & --- \\
\quad All writers, heads only & 0.55 \(\pm\) 0.07 & 0.35 \(\pm\) 0.10 & 0.22 \(\pm\) 0.15 & --- & --- \\
\quad All writers, heads clamped & 0.58 \(\pm\) 0.07 & 0.11 \(\pm\) 0.04 & \(-\)0.09 \(\pm\) 0.12 & --- & --- \\
\quad Residual, heads only & 0.67 \(\pm\) 0.04 & 0.70 \(\pm\) 0.05 & 0.33 \(\pm\) 0.14 & --- & --- \\
\quad Residual, heads clamped & 0.51 \(\pm\) 0.07 & 0.08 \(\pm\) 0.04 & 0.01 \(\pm\) 0.10 & --- & --- \\
\midrule
\multicolumn{6}{l}{\textit{Granite-H}: reader L35, 8 of 32 heads, gate 78/80\%} \\
Writer L24 swapped & 0.49 \(\pm\) 0.07 & 0.20 \(\pm\) 0.09 & 0.18 \(\pm\) 0.23 & 1.01 \(\pm\) 0.02 & 0.21 \(\pm\) 0.13 \\
Control L27 swapped & \(-\)0.01 \(\pm\) 0.01 & 0.00 \(\pm\) 0.01 & 1.02 \(\pm\) 0.02 & 1.01 \(\pm\) 0.01 & 0.00 \(\pm\) 0.01 \\
All writers L24/L26/L29 swapped & 0.52 \(\pm\) 0.07 & 0.17 \(\pm\) 0.12 & 0.08 \(\pm\) 0.27 & 1.01 \(\pm\) 0.02 & 0.21 \(\pm\) 0.16 \\
Control set L27/L31/L33 swapped & 0.02 \(\pm\) 0.01 & 0.02 \(\pm\) 0.02 & 0.99 \(\pm\) 0.03 & 1.01 \(\pm\) 0.02 & 0.04 \(\pm\) 0.02 \\
Residual after L24 replaced & 0.57 \(\pm\) 0.08 & 0.22 \(\pm\) 0.11 & 0.02 \(\pm\) 0.27 & 1.01 \(\pm\) 0.02 & 0.19 \(\pm\) 0.16 \\
\quad Writer swap, heads only & 0.33 \(\pm\) 0.08 & 0.07 \(\pm\) 0.06 & 0.38 \(\pm\) 0.13 & --- & --- \\
\quad Writer swap, heads clamped & 0.61 \(\pm\) 0.09 & 0.14 \(\pm\) 0.07 & \(-\)0.16 \(\pm\) 0.20 & --- & --- \\
\quad All writers, heads only & 0.35 \(\pm\) 0.09 & 0.05 \(\pm\) 0.07 & 0.30 \(\pm\) 0.15 & --- & --- \\
\quad All writers, heads clamped & 0.63 \(\pm\) 0.10 & 0.13 \(\pm\) 0.08 & \(-\)0.21 \(\pm\) 0.21 & --- & --- \\
\quad Residual, heads only & 0.40 \(\pm\) 0.08 & 0.08 \(\pm\) 0.07 & 0.24 \(\pm\) 0.14 & --- & --- \\
\quad Residual, heads clamped & 0.63 \(\pm\) 0.09 & 0.14 \(\pm\) 0.07 & \(-\)0.19 \(\pm\) 0.20 & --- & --- \\
\midrule
\multicolumn{6}{l}{\textit{RecurrentGemma-9B}: reader L29, 16 of 16 heads, gate 80/75\%} \\
Writer L28 swapped & 0.34 \(\pm\) 0.04 & 0.39 \(\pm\) 0.06 & 0.70 \(\pm\) 0.15 & 1.00 \(\pm\) 0.02 & 0.29 \(\pm\) 0.08 \\
Control L22 swapped & 0.01 \(\pm\) 0.01 & 0.00 \(\pm\) 0.01 & 0.98 \(\pm\) 0.03 & 1.01 \(\pm\) 0.02 & 0.00 \(\pm\) 0.01 \\
Residual after L28 replaced & \multicolumn{5}{l}{Equals the guest transplant (the writer immediately precedes the reader)} \\
\quad Writer swap, heads only & 0.36 \(\pm\) 0.07 & 0.36 \(\pm\) 0.08 & 0.61 \(\pm\) 0.14 & --- & --- \\
\quad Writer swap, heads clamped & 0.45 \(\pm\) 0.11 & 0.18 \(\pm\) 0.04 & 0.27 \(\pm\) 0.10 & --- & --- \\
\midrule
\multicolumn{6}{l}{\textit{Nemotron-3-Nano-30B}: reader L33, 32 of 32 heads, gate 82/78\%} \\
Writer L30 swapped & 0.08 \(\pm\) 0.03 & 0.11 \(\pm\) 0.04 & 0.94 \(\pm\) 0.05 & 1.01 \(\pm\) 0.02 & 0.09 \(\pm\) 0.03 \\
Control L21 swapped & \(-\)0.01 \(\pm\) 0.01 & 0.00 \(\pm\) 0.01 & 1.01 \(\pm\) 0.03 & 1.00 \(\pm\) 0.01 & 0.00 \(\pm\) 0.01 \\
All writers L25/L28/L30/L32 swapped & 0.64 \(\pm\) 0.04 & 0.51 \(\pm\) 0.06 & 0.30 \(\pm\) 0.14 & 0.99 \(\pm\) 0.03 & 0.53 \(\pm\) 0.11 \\
Control set L14/L16/L18/L21 swapped & 0.01 \(\pm\) 0.03 & 0.02 \(\pm\) 0.03 & 1.00 \(\pm\) 0.05 & 1.00 \(\pm\) 0.02 & 0.00 \(\pm\) 0.03 \\
Residual after L30 replaced & 0.85 \(\pm\) 0.02 & 0.88 \(\pm\) 0.03 & 0.28 \(\pm\) 0.14 & 1.00 \(\pm\) 0.03 & 0.97 \(\pm\) 0.16 \\
\quad Writer swap, heads only & 0.18 \(\pm\) 0.04 & 0.06 \(\pm\) 0.03 & 0.73 \(\pm\) 0.08 & --- & --- \\
\quad Writer swap, heads clamped & 0.38 \(\pm\) 0.06 & 0.05 \(\pm\) 0.02 & 0.36 \(\pm\) 0.06 & --- & --- \\
\quad All writers, heads only & 0.65 \(\pm\) 0.04 & 0.43 \(\pm\) 0.07 & 0.22 \(\pm\) 0.12 & --- & --- \\
\quad All writers, heads clamped & 0.55 \(\pm\) 0.07 & 0.18 \(\pm\) 0.04 & 0.16 \(\pm\) 0.07 & --- & --- \\
\quad Residual, heads only & 0.78 \(\pm\) 0.04 & 0.65 \(\pm\) 0.05 & 0.19 \(\pm\) 0.11 & --- & --- \\
\quad Residual, heads clamped & 0.66 \(\pm\) 0.07 & 0.37 \(\pm\) 0.04 & 0.15 \(\pm\) 0.07 & --- & --- \\
\bottomrule
\end{tabular}
}
\caption{Writer swap.  Redirection of the recipient's lookup from the
host's key to the guest's, which normalizes the guest-minus-host contrast
between the two transplants (0 = host transplant, 1 = guest transplant),
the guest and host lookups \(g\) and \(h\) of
Equation~\ref{eq:writer-swap}, the
shared-key lookup retained under answer-swap guests, and the host's own
logit shift toward the guest key's value; 95\% identity-bootstrap intervals
over 200 identities.}
\label{tab:writer-swap}
\end{table}

\paragraph{Which recurrent outputs carry the key.}
Table~\ref{tab:writer-singles} reports a prespecified follow-up on the same
pairs: each nominated writer and each matched control swapped alone, each
leave-one-out set, and the full sets, classified by two rules fixed before
the run (\emph{concentrated}: one writer carries at least half of the joint
effect and removing it drops the set below half; \emph{distributed}: every
single writer below half and every leave-one-out set below the joint set).
Granite-H meets the concentrated rule at L24.  Nemotron-H meets the
distributed rule, and its low-increment layers also create guest lookup,
consistent with a key distributed over the recurrent layers.

\begin{table}[htbp]
\centering
\scriptsize
\setlength{\tabcolsep}{4pt}
\begin{tabular}{lrrl}
\toprule
Swap & Guest lookup & Host lookup & Paired guest-lookup difference \\
\midrule
\multicolumn{4}{l}{\textit{Nemotron-H-8B}: writers L28/L33/L39, controls L31/L35/L37, reader L40; pattern: distributed} \\
L28 alone & 0.12 \(\pm\) 0.04 & 0.78 \(\pm\) 0.08 & vs.\ L31: 0.16 \(\pm\) 0.05 \\
\quad Control L31 alone & \(-\)0.04 \(\pm\) 0.02 & 0.98 \(\pm\) 0.04 & \\
L33 alone & 0.04 \(\pm\) 0.04 & 0.95 \(\pm\) 0.06 & vs.\ L35: \(-\)0.10 \(\pm\) 0.05 \\
\quad Control L35 alone & 0.14 \(\pm\) 0.02 & 0.87 \(\pm\) 0.05 & \\
L39 alone & 0.14 \(\pm\) 0.03 & 0.86 \(\pm\) 0.04 & vs.\ L37: \(-\)0.03 \(\pm\) 0.03 \\
\quad Control L37 alone & 0.17 \(\pm\) 0.02 & 0.87 \(\pm\) 0.05 & \\
All but L28 & 0.19 \(\pm\) 0.05 & 0.79 \(\pm\) 0.09 & Joint minus set: 0.12 \(\pm\) 0.09 \\
All but L33 & 0.24 \(\pm\) 0.06 & 0.64 \(\pm\) 0.10 & Joint minus set: 0.07 \(\pm\) 0.06 \\
All but L39 & 0.17 \(\pm\) 0.08 & 0.37 \(\pm\) 0.14 & Joint minus set: 0.14 \(\pm\) 0.03 \\
All writers & 0.31 \(\pm\) 0.09 & 0.10 \(\pm\) 0.20 & \\
All controls & 0.26 \(\pm\) 0.03 & 0.72 \(\pm\) 0.09 & \\
\midrule
\multicolumn{4}{l}{\textit{Granite-H}: writers L24/L26/L29, controls L27/L31/L33, reader L35; pattern: concentrated} \\
L24 alone & 0.20 \(\pm\) 0.09 & 0.18 \(\pm\) 0.24 & vs.\ L27: 0.20 \(\pm\) 0.09 \\
\quad Control L27 alone & 0.00 \(\pm\) 0.01 & 1.02 \(\pm\) 0.02 & \\
L26 alone & \(-\)0.05 \(\pm\) 0.01 & 1.06 \(\pm\) 0.05 & vs.\ L31: \(-\)0.04 \(\pm\) 0.02 \\
\quad Control L31 alone & \(-\)0.01 \(\pm\) 0.01 & 1.00 \(\pm\) 0.01 & \\
L29 alone & 0.02 \(\pm\) 0.01 & 1.02 \(\pm\) 0.04 & vs.\ L33: \(-\)0.01 \(\pm\) 0.02 \\
\quad Control L33 alone & 0.03 \(\pm\) 0.01 & 0.97 \(\pm\) 0.03 & \\
All but L24 & \(-\)0.03 \(\pm\) 0.02 & 1.08 \(\pm\) 0.06 & Joint minus set: 0.20 \(\pm\) 0.11 \\
All but L26 & 0.24 \(\pm\) 0.11 & 0.09 \(\pm\) 0.26 & Joint minus set: \(-\)0.07 \(\pm\) 0.02 \\
All but L29 & 0.15 \(\pm\) 0.11 & 0.15 \(\pm\) 0.25 & Joint minus set: 0.03 \(\pm\) 0.03 \\
All writers & 0.17 \(\pm\) 0.12 & 0.08 \(\pm\) 0.27 & \\
All controls & 0.02 \(\pm\) 0.02 & 0.99 \(\pm\) 0.03 & \\
\bottomrule
\end{tabular}

\caption{Single-writer and leave-one-out swaps on the writer-swap pairs.
Guest and host lookups as fractions of the guest and host transplants'
(95\% identity-bootstrap intervals); the last column gives the paired
guest-lookup difference used by the classification rules.}
\label{tab:writer-singles}
\end{table}

\begin{figure}[htbp]
  \centering
  \includegraphics[width=\linewidth]{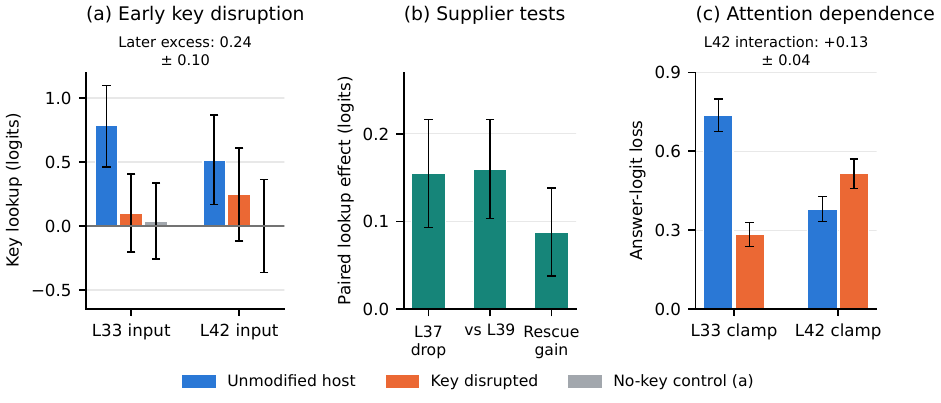}
  \caption{\textbf{A recurrent supplier contributes to Nano's later lookup.}
  (a) Host-key-vs-reference lookup at the two attention-layer inputs; the paired
  later excess over the no-key control is positive.
  (b) Paired supplier interruption, specificity, and rescue effects.
  (c) Answer-logit loss under head clamps before and after disruption;
  the annotated interaction tests increased L42 dependence.
  Intervals resample identities (95\%).}
  \label{fig:delayed-key}
\end{figure}

\paragraph{Delaying the key.}
The no-key guest is the host prompt with its first-hop value replaced by a
dead-end key.  Its outputs at every key-writing layer before the
resolving layer replace the host's, the run continues, and the resolver's
and the next attention layer's inputs are transplanted into the recipient.
The delay strongly attenuates the lookup at the resolver's input in Nano
and RecurrentGemma, and in Nano leaves a lookup at the later layer's input
that the later layer's heads carry (Figure~\ref{fig:delayed-key}a).

\paragraph{Interrupting and rescuing Nano's late supplier.}
The prespecified follow-up uses the delay experiment's Nano identities, early
disruption, and reader sites.  The profile nominates recurrent layer L37,
with L39 as the comparison layer in the same span.  We replace either
layer's final-token output with its no-key donor output, or restore L37's
undisturbed host output for rescue, and transplant the resulting pre-L42
state into the recipient.  All four frozen questions are supported
(Table~\ref{tab:late-supply}): interruption reduces the later lookup by
0.154 \(\pm\) 0.061 where the comparison layer costs nothing, rescue
recovers 0.087 \(\pm\) 0.050 of that and leaves the lookup within the
third of the uninterrupted run that the rule fixed in advance allows,
0.067 \(\pm\) 0.039 below the uninterrupted
run, and clamping the later heads costs more after disruption than in the
undisturbed host.

\FloatBarrier   

\begin{table}[htbp]
\centering
\scriptsize
\begin{tabular}{ll}
\toprule
Paired contrast & Estimate (95\%) \\
\midrule
Later excess over no-key & 0.242 \(\pm\) 0.097 \\
Q1: L37 interruption drop & 0.154 \(\pm\) 0.061 \\
L39 comparison drop & \(-\)0.005 \(\pm\) 0.017 \\
Q2: L37 versus L39 drop & 0.159 \(\pm\) 0.057 \\
Q3: rescue improvement & 0.087 \(\pm\) 0.050 \\
Rescue shortfall from uninterrupted & 0.067 \(\pm\) 0.039 \\
L37/L41 joint interruption drop (secondary) & 0.222 \(\pm\) 0.072 \\
L42 clamp loss, undisturbed host & 0.379 \(\pm\) 0.047 \\
L42 clamp loss, disrupted host & 0.514 \(\pm\) 0.056 \\
Q4: increase in L42 clamp loss & 0.134 \(\pm\) 0.037 \\
L33 clamp loss, undisturbed host & 0.736 \(\pm\) 0.061 \\
L33 clamp loss, disrupted host & 0.282 \(\pm\) 0.046 \\
L37 interruption, share of the surviving lookup & 64\% [48, 85] \\
L39 comparison, share of the surviving lookup & \(-\)2\% [\(-\)12, 4] \\
Rescue, share of the L37 interruption drop & 57\% [31, 79] \\
\bottomrule
\end{tabular}

\caption{Nano late-supplier follow-up on the original delay pool.
Recipient key lookup is the host-key-vs-reference logit-change contrast;
host clamp loss is the change in the host answer's logit.  All entries
are paired effects with 95\% identity-bootstrap intervals.  Q3 also has
a separately frozen point-tolerance criterion, described in the text.  The
last three rows are the ratios the main text quotes, resampled as ratios,
so their intervals are asymmetric.}
\label{tab:late-supply}
\end{table}

\FloatBarrier
\section{The attention reader}
\label{app:reader}

\paragraph{Nominating the site and the heads.}
Sites and heads were nominated by activation patching on separate
discovery examples: a clean activation is restored into a run whose target
value is corrupted, and the top-ranked heads at a layer are tested on fresh
examples against random head sets of the same size.  The site and the
top-eight head set were fixed before any crossed experiment.

\paragraph{Path patching details.}
Under \(F\) we cache the output of every head at the resolving layer, and
starting from the unmodified recipient we inject only the head-output
changes produced under \(F\) for a chosen set \(T\),
\begin{equation}
  o_h\;\leftarrow\;o_h+\mathbf{1}[h\in T]\,\bigl(o_h^{F}-o_h^{R}\bigr),
  \label{eq:path}
\end{equation}
where \(o_h^{F}\) and \(o_h^{R}\) are head \(h\)'s outputs under the
transplant and in the unmodified recipient.  Two clamps complete the
design: a residual bypass \(R_{\rm by}\) holds every head at its recipient
output, and a donor-heads control \(D_{\rm attn}\) holds them at the output
they produced in the donor prompt; \(R_{\rm by}\) plus the all-head
injection reproduces \(F\) exactly.  Each head set is compared with a bank
of random sets of the same size.

\begin{figure}[ht]
 \centering
 \includegraphics[width=\linewidth]{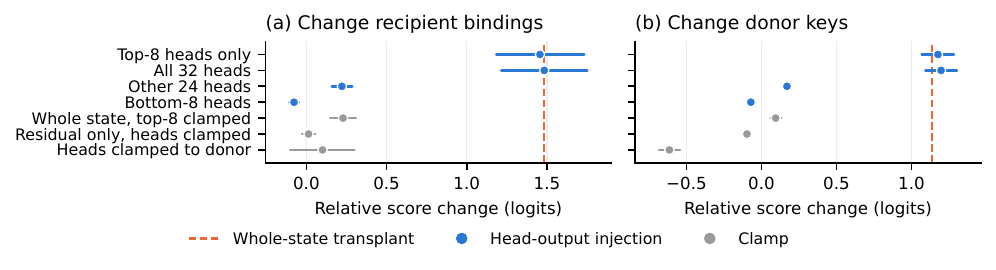}
 \caption{\textbf{Eight attention heads transmit most of the lookup effect.}
 Nemotron-H, L40, 200 identities, 95\% intervals.
 The two panels change recipient bindings or donor keys, respectively;
 selectivity is the predicted value's score increase minus the mean increase
 of the other two recipient-bound values.  Dashed lines show the whole-state
 transplant.  Injecting the eight selected heads' changes nearly reproduces
 it; holding their outputs fixed sharply reduces it.
 Detailed values and Granite-H's comparison are in
 Appendix Tables~\ref{tab:route} and~\ref{tab:route-granite}.}
 \label{fig:route}
\end{figure}

\begin{table}[t]
\centering
\scriptsize
\setlength{\tabcolsep}{9pt}
\renewcommand{\arraystretch}{1}
\begin{tabular}{llrrr}
\toprule
Condition & What carries the transplant & Mapping selectivity & Key selectivity & Top-1 (\%) \\
\midrule
\(F\) & Whole state & 1.48 \(\pm\) 0.29 & 1.14 \(\pm\) 0.10 & 42 \\
\(P_{\rm T}\) & Top-8 head outputs only & 1.46 \(\pm\) 0.28 & 1.18 \(\pm\) 0.10 & 40 \\
\(P_{\rm all}\) & All 32 head outputs & 1.48 \(\pm\) 0.27 & 1.20 \(\pm\) 0.10 & 41 \\
\(P_{\bar{\rm T}}\) & Other 24 head outputs & 0.22 \(\pm\) 0.06 & 0.17 \(\pm\) 0.03 & 4 \\
\(P_{\rm B}\) & Bottom-8 head outputs & \(-\)0.08 \(\pm\) 0.03 & \(-\)0.07 \(\pm\) 0.01 & 3 \\
\(R_{\rm by}\) & Residual; heads clamped to recipient & 0.01 \(\pm\) 0.04 & \(-\)0.10 \(\pm\) 0.02 & 3 \\
\(D_{\rm attn}\) & Residual; heads clamped to donor & 0.10 \(\pm\) 0.20 & \(-\)0.61 \(\pm\) 0.07 & 2 \\
\(C_{\rm T}\) & Whole state, top-8 clamped & 0.23 \(\pm\) 0.08 & 0.09 \(\pm\) 0.04 & 3 \\
\bottomrule
\end{tabular}
\caption{Path patching at L40 in Nemotron-H on the common assay (200
identities), in logits with 95\% intervals.  Top-1 is the seven-way rate
for the predicted candidate; the unmodified recipient scores 4\%.
\(R_{\rm by}+P_{\rm all}\) reproduces \(F\) to zero error.}
\label{tab:route}
\end{table}

\begin{table}[htbp]
\centering
\scriptsize
\setlength{\tabcolsep}{4pt}
\begin{tabular}{llrrr}
\toprule
Condition & What carries the transplant & Mapping selectivity & Key selectivity & Top-1 (\%) \\
\midrule
\(F\) & Whole state & 1.44 \(\pm\) 0.29 & 1.09 \(\pm\) 0.10 & 38 \\
\(P_{\rm T}\) & Top-8 head outputs only & 1.00 \(\pm\) 0.20 & 0.81 \(\pm\) 0.08 & 25 \\
\(P_{\rm all}\) & All 32 head outputs & 1.42 \(\pm\) 0.27 & 1.15 \(\pm\) 0.10 & 41 \\
\(P_{\bar{\rm T}}\) & Other 24 head outputs & 0.52 \(\pm\) 0.11 & 0.41 \(\pm\) 0.05 & 8 \\
\(P_{\rm B}\) & Bottom-8 head outputs & \(-\)0.14 \(\pm\) 0.04 & \(-\)0.12 \(\pm\) 0.02 & 4 \\
\(R_{\rm by}\) & Residual; heads clamped to recipient & 0.04 \(\pm\) 0.06 & \(-\)0.07 \(\pm\) 0.04 & 3 \\
\(D_{\rm attn}\) & Residual; heads clamped to donor & 0.06 \(\pm\) 0.18 & \(-\)0.36 \(\pm\) 0.08 & 3 \\
\(C_{\rm T}\) & Whole state, top-8 clamped & 0.57 \(\pm\) 0.14 & 0.35 \(\pm\) 0.06 & 8 \\
\bottomrule
\end{tabular}
\caption{Path patching at L35 in Granite-H on the common assay, in the
format of Table~\ref{tab:route}.  Head sets were ranked on a 30-identity
discovery pool and frozen before any test identity; the random bank holds
255 eight-head sets.  The top-eight route carries 69\% \(\pm\) 6 of
\(F\)'s mapping and 75\% \(\pm\) 4 of its key selectivity, the rest
spread across the other heads; it beats the bottom eight by 1.14 and 0.93
logits and every random set (\(p=1/256\)); \(R_{\rm by}+P_{\rm all}\)
reproduces \(F\) to zero error.}
\label{tab:route-granite}
\end{table}

\begin{table}[htbp]
\centering
\scriptsize
\setlength{\tabcolsep}{4pt}
\begin{tabular}{llrrr}
\toprule
Condition & What carries the transplant & Mapping selectivity & Key selectivity & Top-1 (\%) \\
\midrule
\(F\) & Whole state & 1.18 \(\pm\) 0.25 & 1.11 \(\pm\) 0.13 & 28 \\
\(P_{\rm T}\) & Top-8 head outputs only & 1.14 \(\pm\) 0.26 & 1.04 \(\pm\) 0.13 & 26 \\
\(P_{\rm all}\) & All 32 head outputs & 1.01 \(\pm\) 0.23 & 0.93 \(\pm\) 0.12 & 18 \\
\(P_{\bar{\rm T}}\) & Other 24 head outputs & \(-\)0.14 \(\pm\) 0.04 & \(-\)0.12 \(\pm\) 0.02 & 3 \\
\(P_{\rm B}\) & Bottom-8 head outputs & \(-\)0.17 \(\pm\) 0.05 & \(-\)0.14 \(\pm\) 0.02 & 3 \\
\(R_{\rm by}\) & Residual; heads clamped to recipient & 0.43 \(\pm\) 0.10 & 0.43 \(\pm\) 0.06 & 4 \\
\(D_{\rm attn}\) & Residual; heads clamped to donor & 0.33 \(\pm\) 0.13 & 0.16 \(\pm\) 0.06 & 2 \\
\(C_{\rm T}\) & Whole state, top-8 clamped & 0.26 \(\pm\) 0.08 & 0.26 \(\pm\) 0.04 & 4 \\
\bottomrule
\end{tabular}
\caption{Path patching at L33 in Nemotron-3 Nano on the common assay, in
the format of Table~\ref{tab:route}.  Head sets were ranked on a
30-identity discovery pool and frozen before any test identity (L33 heads
7, 5, 11, 0, 4, 1, 15, 13).  The top-eight route carries 96\% \(\pm\) 7 of
\(F\)'s mapping and 93\% \(\pm\) 4 of its key selectivity, exceeds the
bottom eight by 1.31 and 1.18 logits, and exceeds every one of 63 random
eight-head sets (\(p=1/64\)).  Because Nano has a second attention layer
after L33 (L42), the residual bypass keeps part of the effect.}
\label{tab:route-nano}
\end{table}

\subsection{Intervention-family comparison}
\label{app:scrub-results}

Causal scrubbing resamples each donor state from the two other cells that
share one factor and differ in the other two, preserving the key ($S_C$),
answer ($S_A$), or row ($S_P$).  Preservation is the negative centered
mean-squared error between the seven-candidate logit changes:
\begin{equation}
  \mathrm{Pres}(S)=-\,\bigl\|\,\widetilde{\Delta\ell}_{S}-\widetilde{\Delta\ell}_{F}\bigr\|_2^2/7.
  \label{eq:scrub}
\end{equation}

In all three hybrids, preserving the key preserves the transplant's
pattern of score changes better than preserving the answer or the row
(Table~\ref{tab:scrub-results}); for each cell, the two other cells that
share exactly the preserved factor supply the resamples.
Table~\ref{tab:prior-manipulations} relates these interventions to earlier
manipulations of in-context binding.

\label{app:method-comparison}

\begin{table}[htbp]
\centering
\small
\setlength{\tabcolsep}{4pt}
\begin{tabular}{lccc rr}
\toprule
 & \multicolumn{3}{c}{Key's answer against the recipient's others (\(\times\))}
 & \multicolumn{2}{c}{Preservation advantage} \\
\cmidrule(lr){2-4}\cmidrule(lr){5-6}
Model & shares the key & shares the answer & shares the row
 & over answer & over row \\
\midrule
Nemotron-H & 3.42 \(\pm\) 0.35 & 0.62 \(\pm\) 0.04 & 0.47 \(\pm\) 0.02
 & 0.72 \(\pm\) 0.11 & 0.86 \(\pm\) 0.11 \\
Granite-H & 3.18 \(\pm\) 0.35 & 0.62 \(\pm\) 0.04 & 0.51 \(\pm\) 0.03
 & 0.97 \(\pm\) 0.12 & 1.08 \(\pm\) 0.11 \\
Nemotron-3 Nano & 3.29 \(\pm\) 0.42 & 0.61 \(\pm\) 0.05 & 0.52 \(\pm\) 0.04
 & 1.14 \(\pm\) 0.15 & 1.12 \(\pm\) 0.14 \\
\bottomrule
\end{tabular}
\caption{\textbf{Preserving the key better preserves the transplant's effect.}
Left: after the donor state is resampled from donors that share the named
factor, the factor by which the key's answer gains over the recipient's
other candidates, the exponentiated contrast reported in
Table~\ref{tab:acts-as-key}, with 95\% intervals; \(\times 1\) is no change.
Right: the preregistered test, the preservation advantages of
key-preserving over answer- and row-preserving resampling
(Eq.~\ref{eq:scrub}).  Positive values indicate a reduction in centered
mean-squared error relative to the original whole-state transplant.}
\label{tab:scrub-results}
\end{table}

\begin{table}[htbp]
\centering
\scriptsize
\setlength{\tabcolsep}{3pt}
\begin{tabular}{
  >{\raggedright\arraybackslash}p{0.17\linewidth}
  >{\raggedright\arraybackslash}p{0.20\linewidth}
  >{\raggedright\arraybackslash}p{0.13\linewidth}
  >{\raggedright\arraybackslash}p{0.42\linewidth}}
\toprule
Family & Unit & Learned d.o.f. & Held-out outcome on the common assay \\
\midrule
Crossed transplant & Complete residual & 0 &
Key-dominant, dual-supported; mapping 1.48, key 1.14 \\
Path patching & Head-output route & 0 (set prespecified) &
Top-8 route carries 98--104\% of \(F\) in Nemotron-H, 93--96\% in Nano and 69--75\% in Granite-H \\
Causal scrubbing & Factor-preserving resample & 0 &
Key-preserving scrub preserves, answer- and row-preserving do not, in all three hybrids \\
\bottomrule
\end{tabular}
\caption{The three families on identical held-out identities.
Method-specific questions differ, so the outcome column is not a ranking.}
\end{table}

\begin{table}[htbp]
\centering
\scriptsize
\setlength{\tabcolsep}{3pt}
\begin{tabular}{
  >{\raggedright\arraybackslash}p{0.15\linewidth}
  >{\raggedright\arraybackslash}p{0.27\linewidth}
  >{\raggedright\arraybackslash}p{0.23\linewidth}
  >{\raggedright\arraybackslash}p{0.28\linewidth}}
\toprule
Study & Manipulation & What it identifies & What crossing adds \\
\midrule
\citet{feng2024how} & Binding-ID vectors added to, or swapped between, entity--attribute pairs & A binding subspace shared across entities and its geometry & A reference baseline and a key \(\times\) position factorial, so that ``binding'' splits into the bound value and the bound row, each with a preregistered bound \\
\citet{gur-arieh2026mixing} & Counterfactual prompt pairs that vary one retrieval cue (positional, lexical, reflexive) with activation patching & Which retrieval mechanism a model uses at a site, and that the mechanisms mix & Computed-key assembly in hybrids compared with dense models; a reference baseline and prespecified factorial classification \\
\citet{sharma2026llms} & Patching a predicate (filter) representation between list prompts & A general filter-head circuit that selects list items & The key and position of a bound variable rather than a predicate, and three intervention families compared on one assay \\
\bottomrule
\end{tabular}
\caption{Relation to prior manipulations of in-context binding.  Earlier binding studies also use counterfactual pairs and whole-state
interventions.  Our comparisons extend that approach to computed keys,
hybrid architectures, and shared outcomes across intervention families.}
\label{tab:prior-manipulations}
\end{table}

\FloatBarrier
\section{Retrieval after the last attention layer}
\label{app:suffix-retrieval}

\subsection{Shifted transplant}

\paragraph{Shifted transplant.}
Moving the donor state one layer past the final attention layer removes the
key effect on the arrow and binding tasks in Nemotron-H and Granite-H,
while Qwen2.5, with attention after the shifted site, keeps it.  A state
inserted one layer later than it was captured is off-distribution by
construction; the fraction of runs whose final-layer state falls in the tail
of natural states stays close to that of unmodified recipients in the
hybrids.  On the word task part of the key effect survives past Nemotron-H's
last attention layer, which motivates Section~\ref{sec:suffix}; its numeric
twin, with room numbers in place of location words, keeps none
(Figure~\ref{fig:suffix-clamp}a).

\subsection{Recurrent outputs after the last attention layer}

\begin{figure}[htbp]
  \centering
  \includegraphics[width=\linewidth]{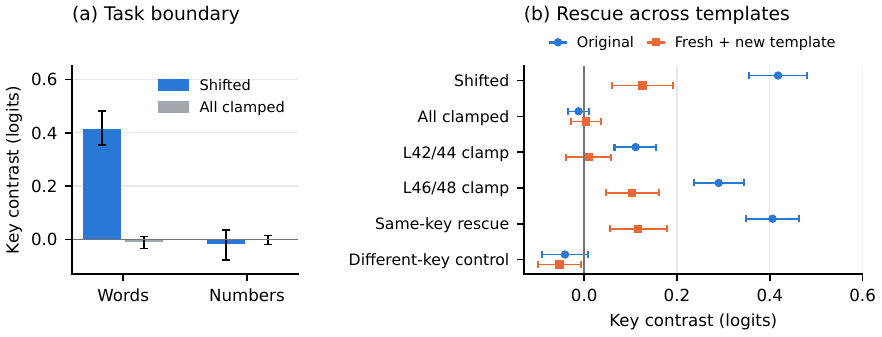}
  \caption{\textbf{Key-specific rescue of the post-attention lookup.}
  Nemotron-H, key effect \(\kappa\) in logits with 95\% identity-bootstrap intervals.
  (a) Shifted lookup and collective Mamba clamp on fresh word/numeric twins.
  (b) The nominated pair, comparison pair, and cross-donor rescue in the
  original word pool and a fresh pool under a new template, with all sites frozen.
  Both support all six primary contrasts.}
  \label{fig:suffix-clamp}
\end{figure}

\paragraph{The post-attention lookup depends on Mamba outputs.}
Two prespecified follow-ups on fresh word-task identities test the shifted
lookup of Nemotron-H (Figure~\ref{fig:suffix-clamp}).  Holding the
final-token outputs of the five post-attention Mamba layers at the
unmodified recipient's values removes the lookup, and numeric twins show
none to remove.  The pair was nominated from the word-task depth profile on an earlier
identity pool, where L42 and L44 carry most of the post-attention rise, and
not from any single-layer result.  Clamping the nominated pair L42/L44 removes more than a
comparison pair, and restoring those outputs from a different donor with the
same key brings the lookup back, while a donor with a different key does
not; all six frozen criteria pass, again on fresh graphs under a new
template (Table~\ref{tab:suffix-rescue}).  Every exact check is zero.
Clamped one at a time, the five layers cost the shifted lookup
\(0.063\), \(\mathbf{0.224}\), \(0.118\), \(0.018\) and \(0.084\)
logits on the first pool and \(0.021\), \(\mathbf{0.094}\), \(0.023\),
\(-0.000\) and \(0.003\) on the fresh one, for L42, L44, L46, L48 and
L50: L44 is the largest in both, and L48 costs nothing in either.

\begin{table}[htbp]
\centering
\scriptsize
\begin{tabular}{lll}
\toprule
Condition or paired contrast & Original (95\%) & Fresh template (95\%) \\
\midrule
Shifted lookup & 0.418 \(\pm\) 0.063 & 0.126 \(\pm\) 0.066 \\
All five outputs clamped & \(-\)0.012 \(\pm\) 0.023 & 0.004 \(\pm\) 0.032 \\
L42/L44 clamped & 0.111 \(\pm\) 0.045 & 0.010 \(\pm\) 0.049 \\
L46/L48 clamped & 0.290 \(\pm\) 0.053 & 0.104 \(\pm\) 0.057 \\
Same-key, different-answer rescue & 0.406 \(\pm\) 0.057 & 0.117 \(\pm\) 0.060 \\
Different-key, same-answer control & \(-\)0.041 \(\pm\) 0.050 & \(-\)0.053 \(\pm\) 0.046 \\
Collective clamp drop & 0.430 \(\pm\) 0.052 & 0.122 \(\pm\) 0.048 \\
Nominated-pair drop & 0.307 \(\pm\) 0.037 & 0.116 \(\pm\) 0.031 \\
Pair specificity & 0.179 \(\pm\) 0.042 & 0.094 \(\pm\) 0.030 \\
Rescue improvement & 0.295 \(\pm\) 0.038 & 0.107 \(\pm\) 0.029 \\
Rescue key specificity & 0.447 \(\pm\) 0.063 & 0.170 \(\pm\) 0.050 \\
\bottomrule
\end{tabular}

\caption{Nemotron-H suffix localization and cross-donor rescue, and their
prospective replication on fresh graphs under a new template.  Key-vs-reference
logit changes and paired differences, with 95\% identity-bootstrap intervals
(200 identities per pool).  Clamps use recipient baselines; rescue/control
use different shifted donors.  All six primary criteria pass in each pool.}
\label{tab:suffix-rescue}
\end{table}

\subsection{Recurrent-memory intervention}
\label{app:memory}

For each identity, three keys \(m_c\) are bound to three places \(B_c\).
For a fixed point \(d\ne c\), a control memory prompt uses the recipient's
facts and asks a two-hop question whose intermediate key is \(m_d\) and
answer \(B_d\); its transposed partner has the same question, answer, row
layout, and values, but exchanges the places bound to the two other keys,
so it binds the queried \(m_c\) to a third value, the \emph{supplied} value.
Every token except the final one is prefilled into the model's cache; at the
final token only the recurrent states of the post-attention layers are
replaced, while the residual entering them and the cached convolution
histories belong to the run being tested.  Swapping all post-attention
memories moves the answer in both conditions, and so do smaller swaps
(Table~\ref{tab:memory-read-detail}); the result replicates under a second
template with every site, endpoint, and rule frozen.  A trajectory audit
covers every condition of the swap, the ordinary-regime ones included, and
finds no excess final-layer atypicality in any of them: against the
unmodified recipient each paired contrast is \(-\)0.01 \(\pm\) 0.01, for the ordinary swaps as for the shifted.  The one
condition it cannot clear is the plain same-site transplant, whose interval
still contains zero.  In every run the single-token decode matches a full forward
within the frozen bars, and every exact check is zero.

\paragraph{Granite-H.}
Granite-H carries two of the paper's targeted experiments but takes no
memory role.  It is not excluded by choice: on the two wordings screened it
answers 53\% and 58\% of donor prompts, 38\% and 46\% of recipients, and
47\% and 57\% of memory prompts, so all three roles fall below the 60\%
gate every model passes before it is assayed
(Section~\ref{sec:setup}).  A replication on this task setup would not be
interpretable for it.

\paragraph{Key specificity.}
A preregistered control repeats the swap with the memory's fixed point set
to the queried key \(m_c\).  Its transposed memory then binds \(m_c\) to
the place the test prompt already binds it to and exchanges the places of
the other two keys instead, so nothing the question asks about moves and a
reader of the installed memory should leave the answer alone.  In
Nemotron-H on 200 identities the two exchanged places gain nothing over the
queried answer under the shifted state: \(-\)0.005 \(\pm\) 0.020 logits
for the all-layer swap, and intervals containing zero for the pair and for
L44 alone.  In ordinary execution a
small residual remains, 0.060 \(\pm\) 0.021 for the all-layer
swap and 0.042 \(\pm\) 0.020 for L44, against \(0.978\) for
the exchange that reaches the queried binding; it sits on one of the two
exchanged places rather than both.  Nano-9B-v2 and Granite-Small repeat the
control on their all-layer swaps with the same shape: the exchanged places
gain nothing under the shifted state, \(-\)0.004 \(\pm\) 0.016 and
0.007 \(\pm\) 0.020, and the same small residual remains under
ordinary execution, 0.038 \(\pm\) 0.013 and 0.034 \(\pm\) 0.016.

\paragraph{Rotated key rows.}
The memories above keep the test prompt's row order, so the place a memory
binds to the queried key also sits in that key's row and the two accounts
predict the same answer.  The crossed memory separates them: it holds the
same exchanged bindings and writes them into key rows rotated by one, each
row taking the bindings of the row below it and the last taking the first's, so
the place bound to the queried key now sits in another key's row, and the
row the queried key occupies holds a third place.  Everything else follows
the swap above.  The control is the same unexchanged memory in the
recipient's own row order, so the crossed memory differs from it in two
ways at once, the exchange and the layout; what separates the two accounts
is the comparison between the two places the crossed memory puts in play,
both read against that one control.  The answer follows the key, under both
incoming states and in almost every identity (Table~\ref{tab:crossed-appendix}).
The rotation costs the read little: against the test prompt's own answer
the supplied place gains \(1.10\) logits when the memory keeps the
recipient's row order and \(0.89\) when its rows are rotated.  The run is
the swap protocol of Section~\ref{sec:suffix} on a fresh pool of 200
identities, with its sites, sample, gates and exact checks unchanged; the
crossed memory and the key-versus-row contrast were fixed in its
configuration before the run.  That configuration fixed the shifted state as
the second incoming state, where Section~\ref{sec:suffix} reports the
skipped one, so the two meet in ordinary execution, which is the column
Table~\ref{tab:crossed-memory} quotes.

\begin{table}[htbp]
\centering
\scriptsize
\setlength{\tabcolsep}{3pt}
\begin{tabular}{llrrrr}
\toprule
Model & Incoming state & Key over row & Key over test prompt's & Row over test prompt's & Identities, key \(>\) row \\
\midrule
Nemotron-H & Ordinary execution & 0.65 \(\pm\) 0.04 & 0.89 \(\pm\) 0.05 & 0.24 \(\pm\) 0.03 & 98\% \\
 & Shifted & 0.82 \(\pm\) 0.05 & 1.00 \(\pm\) 0.07 & 0.18 \(\pm\) 0.03 & 100\% \\
Granite-Small & Ordinary execution & 0.31 \(\pm\) 0.03 & 0.40 \(\pm\) 0.02 & 0.09 \(\pm\) 0.01 & 98\% \\
 & Shifted & 0.38 \(\pm\) 0.03 & 0.48 \(\pm\) 0.03 & 0.10 \(\pm\) 0.01 & 100\% \\
\bottomrule
\end{tabular}

\caption{The crossed memory, which writes the same bindings into key rows
rotated by one.  Contrasts are logit changes of the crossed memory against
the control memory, differenced between two candidates, with 95\%
identity-bootstrap intervals over 200 identities; the last column is the
share of identities in which the place the key is bound to gains more than
the place now in the key's row.}
\label{tab:crossed-appendix}
\end{table}

\begin{table}[!htb]
 \centering\scriptsize
 \setlength{\tabcolsep}{3pt}
 \resizebox{\linewidth}{!}{\begin{tabular}{llcc}
\toprule
Template & Execution & All five memories & L44 memory only \\
\midrule
Original & Shifted
  & $0.331 \pm 0.036\;/\;0.674 \pm 0.065$
  & $0.187 \pm 0.028\;/\;0.369 \pm 0.042$ \\
Original & Ordinary
  & $0.704 \pm 0.045\;/\;0.978 \pm 0.057$
  & $0.385 \pm 0.034\;/\;0.537 \pm 0.041$ \\
\textbf{Fresh wording} & \textbf{Shifted}
  & $\mathbf{0.111} \pm 0.022\;/\;\mathbf{0.233} \pm 0.036$
  & $\mathbf{0.064} \pm 0.013\;/\;\mathbf{0.118} \pm 0.020$ \\
\textbf{Fresh wording} & \textbf{Ordinary}
  & $\mathbf{0.375} \pm 0.027\;/\;\mathbf{0.530} \pm 0.035$
  & $\mathbf{0.202} \pm 0.018\;/\;\mathbf{0.285} \pm 0.021$ \\
\bottomrule
\end{tabular}
}
 \caption{Recurrent-memory effects for all five layers and L44 alone.
 Each cell gives the supplied value over the memory prompt's own answer /
 over the recipient's value, in logits with 95\% intervals.
 Bold marks the fresh-wording replication.  The all-five-layer results
 also appear with separate contrast columns in Table~\ref{tab:memory-map}.}
 \label{tab:memory-read-detail}
\end{table}

\begin{table}[!htb]
\centering
\scriptsize
\setlength{\tabcolsep}{3.5pt}
\resizebox{\textwidth}{!}{%
\begin{tabular}{llccccccl}
\toprule
 & & Accuracy & \multicolumn{2}{c}{\(\sigma_{\rm own}\): over the memory prompt's answer} & \multicolumn{2}{c}{\(\sigma_{\rm test}\): over the test prompt's answer} & Consistent & \\
\cmidrule(lr){4-5}\cmidrule(lr){6-7}
Model (recurrent layers after last attention) & Task version & (S / R / M, \%) & Shifted & Ordinary & Shifted & Ordinary & identities (\%) & Read \\
\midrule
Nemotron-H-8B (5 Mamba-2) & Word & 71 / 68 / 69 & 0.33 \(\pm\) 0.04 & 0.70 \(\pm\) 0.04 & 0.67 \(\pm\) 0.07 & 0.98 \(\pm\) 0.06 & 100 & Both conditions \\
 & Story & 64 / 61 / 63 & 0.11 \(\pm\) 0.02 & 0.37 \(\pm\) 0.03 & 0.23 \(\pm\) 0.04 & 0.53 \(\pm\) 0.03 & 99 & Both conditions \\
 & Paragraph & 71 / 66 / 70 & 0.44 \(\pm\) 0.04 & 0.79 \(\pm\) 0.05 & 0.89 \(\pm\) 0.09 & 1.09 \(\pm\) 0.06 & 100 & Both conditions \\
 & Short word & 76 / 74 / 78 & 0.33 \(\pm\) 0.04 & 0.78 \(\pm\) 0.04 & 0.68 \(\pm\) 0.08 & 1.05 \(\pm\) 0.06 & 99 & Both conditions \\
 & Numeric twin & 88 / 87 / 86 & 0.00 \(\pm\) 0.01 & 0.00 \(\pm\) 0.00 & 0.00 \(\pm\) 0.01 & 0.00 \(\pm\) 0.01 & 18 & Not detected \\
 & Numeric, one token & 91 / 88 / 87 & 0.01 \(\pm\) 0.01\(^\dagger\) & 0.01 \(\pm\) 0.01 & 0.02 \(\pm\) 0.01\(^\dagger\) & 0.01 \(\pm\) 0.01 & 33 & Ordinary \\
 & Number words & 85 / 82 / 84 & 0.02 \(\pm\) 0.00\(^\dagger\) & 0.03 \(\pm\) 0.00 & 0.05 \(\pm\) 0.01\(^\dagger\) & 0.04 \(\pm\) 0.01 & 60 & Ordinary \\
\addlinespace[2pt]
Nemotron-3 Nano 30B (4 Mamba-2) & Word & 86 / 84 / 87 & 0.13 \(\pm\) 0.02 & 0.20 \(\pm\) 0.03 & 0.27 \(\pm\) 0.03 & 0.27 \(\pm\) 0.03 & 91 & Both conditions \\
 & Story & 75 / 69 / 72 & 0.08 \(\pm\) 0.02\(^\dagger\) & 0.15 \(\pm\) 0.02 & 0.13 \(\pm\) 0.02\(^\dagger\) & 0.21 \(\pm\) 0.03 & 86 & Ordinary \\
 & Paragraph & 69 / 62 / 68 & 0.07 \(\pm\) 0.02 & 0.09 \(\pm\) 0.01 & 0.15 \(\pm\) 0.01 & 0.13 \(\pm\) 0.02 & 76 & Both conditions \\
 & Short word & 86 / 85 / 89 & 0.13 \(\pm\) 0.02 & 0.21 \(\pm\) 0.02 & 0.23 \(\pm\) 0.03 & 0.27 \(\pm\) 0.03 & 88 & Both conditions \\
\addlinespace[2pt]
Nemotron-Nano-9B-v2 (8 Mamba-2) & Word & 88 / 86 / 86 & 0.18 \(\pm\) 0.02 & 0.34 \(\pm\) 0.02 & 0.37 \(\pm\) 0.04 & 0.41 \(\pm\) 0.03 & 99 & Both conditions \\
 & Paragraph & 81 / 76 / 81 & 0.17 \(\pm\) 0.02 & 0.31 \(\pm\) 0.03 & 0.37 \(\pm\) 0.03 & 0.41 \(\pm\) 0.02 & 97 & Both conditions \\
 & Short word & 90 / 88 / 90 & 0.15 \(\pm\) 0.03 & 0.34 \(\pm\) 0.03 & 0.29 \(\pm\) 0.04 & 0.41 \(\pm\) 0.03 & 98 & Both conditions \\
\addlinespace[2pt]
Nemotron-Nano-12B-v2 (4 Mamba-2) & Word & 90 / 86 / 93 & 0.01 \(\pm\) 0.01 & 0.00 \(\pm\) 0.01 & 0.01 \(\pm\) 0.01 & 0.00 \(\pm\) 0.01 & 27 & Not detected \\
 & Short word & 90 / 92 / 92 & 0.00 \(\pm\) 0.01 & 0.00 \(\pm\) 0.01 & 0.01 \(\pm\) 0.01 & 0.00 \(\pm\) 0.01 & 32 & Not detected \\
\addlinespace[2pt]
Nemotron-H-47B (5 Mamba-2) & Word & 96 / 94 / 96 & 0.00 \(\pm\) 0.00 & 0.00 \(\pm\) 0.00 & 0.00 \(\pm\) 0.00 & 0.00 \(\pm\) 0.00 & 27 & Not detected \\
 & Short word & 97 / 98 / 97 & 0.00 \(\pm\) 0.00 & 0.00 \(\pm\) 0.00 & 0.00 \(\pm\) 0.00 & 0.00 \(\pm\) 0.00 & 31 & Not detected \\
\addlinespace[2pt]
Granite-4.0-H-Small (4 Mamba-2) & Word & 83 / 78 / 81 & 0.24 \(\pm\) 0.03 & 0.39 \(\pm\) 0.03 & 0.48 \(\pm\) 0.04 & 0.51 \(\pm\) 0.03 & 99 & Both conditions \\
 & Paragraph & 76 / 72 / 75 & 0.21 \(\pm\) 0.03 & 0.35 \(\pm\) 0.03 & 0.43 \(\pm\) 0.03 & 0.46 \(\pm\) 0.03 & 96 & Both conditions \\
 & Short word & 87 / 82 / 87 & 0.25 \(\pm\) 0.03 & 0.44 \(\pm\) 0.03 & 0.47 \(\pm\) 0.04 & 0.54 \(\pm\) 0.02 & 100 & Both conditions \\
 & Numeric twin & 96 / 95 / 94 & 0.07 \(\pm\) 0.01 & 0.06 \(\pm\) 0.01 & 0.13 \(\pm\) 0.01 & 0.07 \(\pm\) 0.00 & 88 & Both conditions \\
 & Numeric, one token & 96 / 96 / 94 & 0.09 \(\pm\) 0.01 & 0.05 \(\pm\) 0.00 & 0.19 \(\pm\) 0.01 & 0.07 \(\pm\) 0.00 & 80 & Both conditions \\
 & Numeric, two tokens & 94 / 89 / 93 & 0.01 \(\pm\) 0.01\(^\dagger\) & 0.02 \(\pm\) 0.01 & 0.04 \(\pm\) 0.01\(^\dagger\) & 0.02 \(\pm\) 0.01 & 62 & Ordinary \\
\addlinespace[2pt]
RecurrentGemma-9B (2 RG-LRU) & Short word & 70 / 68 / 68 & 0.00 \(\pm\) 0.00 & 0.00 \(\pm\) 0.00 & 0.00 \(\pm\) 0.00 & 0.00 \(\pm\) 0.00 & 40 & Not detected \\
 & Short word, present tense & 84 / 84 / 84 & 0.00 \(\pm\) 0.00 & 0.00 \(\pm\) 0.00 & 0.00 \(\pm\) 0.00 & 0.00 \(\pm\) 0.00 & 46 & Not detected \\
\bottomrule
\end{tabular}
}
\caption{Every matched-memory swap that was run.  All recurrent memories
after the last attention layer are replaced; 200 fresh identities per row;
accuracy is unmodified seven-candidate top-1 for the source, recipient, and
memory prompts (S / R / M).  Scores are \(\sigma_{\rm own}\) and
\(\sigma_{\rm test}\) (Equation~\ref{eq:memory-read}) in logits;
\(e^{\sigma}\) is the geometric-mean probability-ratio factor of
Figure~\ref{fig:memory-read}.
``Consistent identities'' is the share of identities whose ordinary
\(\sigma_{\rm own}\) and \(\sigma_{\rm test}\) are both positive.  ``Read''
applies one rule to every row: both scores have a 95\% lower bound above
zero in that condition.  The two conditions here are ordinary execution and
the shifted state, which is the pair this panel was preregistered under;
Section~\ref{sec:suffix} reports the skipped state as its second condition,
and the two meet in the ordinary column.  \(^\dagger\)The shifted state failed its validity
check, so that condition is not scored.  Task versions: ``Story'' uses a
second sentence template, ``Paragraph'' writes the facts as one paragraph,
``Short word'' is the word task without its three distractor facts, and
``present tense'' states its facts in the present tense.  Checkpoints and revisions are
in Table~\ref{tab:model-panel}; RecurrentGemma-9B runs in float32 for this
experiment.}
\label{tab:memory-map}
\end{table}

\paragraph{Further hybrids and the numeric twin.}
Nemotron-3-Nano-30B (four post-attention memories after attention L42),
Nemotron-Nano-9B-v2 (eight), and Granite-4.0-H-Small (four, after
attention L35) repeat the design on fresh identities with every site,
endpoint, and rule fixed before the run (Table~\ref{tab:memory-map}).  The
models where the swap finds nothing pass the same exact checks as the
others, so the flat result is a measurement and not a failed install: in
Nemotron-H-47B and Nano-12B-v2 the null, self-swap, installed-state,
effective-input and stored-state checks are all exactly zero, and the
single-token decode agrees with the full forward on 1,797 and 1,798 of the
1,800 comparisons.
The carried-answer control is zero in Nemotron-H, Nano, and Granite-Small
and small and positive in Nano-9B-v2 (0.03 in ordinary execution on the
word task); both scores are measured net of it.  Granite-Small's bf16
weights are widened to float32 as each module runs, so that the
single-token decode matches the full forward.  On Nemotron-H's numeric
twin, solved as reliably as the word task, the swap leaves the logits
unchanged.  The twin differs from the word task in two ways at once: its
values are numbers, and each two-digit room number takes three tokens where
every one of the twenty-one place words takes one, so the answer is written
across three positions rather than one.  Separating the two calls for a
value that is numeric and occupies one position.

Three versions do that.  Writing rooms as \texttt{\#4} makes the value one
token after a shared \texttt{\#}, so the answer is read at the position the
swap acts on, as it is on the word task.  Naming rooms with number words
keeps one token per value but makes the value a word again, so it bounds
the two accounts rather than separating them.  The third holds the model
fixed instead of the value: Granite-Small's tokenizer groups digits in
threes, so in that one model \texttt{\#40} is one token after the shared
prefix and \texttt{\#4320} is two, with the template, the value type and
the model unchanged between them.  That pair is the contrast the
word-versus-number twin cannot make, since it varies the value's type and
its length together.

All three ran, each on a fresh pool with its sites, gates and exact checks
unchanged, and Granite-Small's own numeric twin ran beside its pair
(Table~\ref{tab:memory-map}).  Length is not what the twin was measuring.
In Nemotron-H the one-token numeric value leaves the read where the
three-token one left it, undetectable beside the word task.  Granite-Small,
which reads its numeric twin weakly rather than not at all, gains as much
from \texttt{\#40} as from that twin and less from \texttt{\#4320}, so
within one model and one template the read does not follow how many
positions the answer occupies.  Number words, one token and a word, move
further than either numeric version in both models and still fall well
short of places.  What a value must be for these layers to read it is
therefore open, and it is not the pair of differences the
word-versus-number twin puts in play.

The cell we cannot fill is a place word written across several tokens.  The
assay requires the candidates of an identity to share a token prefix and
differ at one position after it, which two-digit room numbers satisfy and
multi-token place words do not: the place words that split in these
tokenizers differ in their first token, not their last, and no family of
ten shares a first token.

\paragraph{A third incoming state.}
Section~\ref{sec:suffix} runs the swap under ordinary execution and under
the skipped state.  A third state removes the test prompt's lead entirely:
the final token's residual entering \(S\) is taken from another prompt, so
nothing the test prompt computed at that token reaches these layers.  What
remains of the test prompt in \(S\) is its convolution window and its
gates; the memory prompt supplies the stored state alone.  Both runs of the
contrast see the same foreign residual, so whatever that residual carries
cancels between the transposed memory and the control, and only the two
exchanged bindings differ.  A preregistered validity condition checks that
the state still performs a key-based lookup rather than producing noise:
with the foreign residual installed and no memory swap, the donor's key
still raises its own place over a candidate that neither account predicts,
by 0.37 \(\pm\) 0.06 logits on the word task and 0.49 \(\pm\) 0.07 on its
paragraph version in Nemotron-H.  The read is
weaker under this state than under ordinary execution, which is why
Section~\ref{sec:suffix} does not use it.

\paragraph{Why the ordinary read exceeds the shifted read.}
Table~\ref{tab:bypass-source} compares the three incoming states under the
same memory swap: \emph{ordinary}
execution, the \emph{skipped} state, which is the test prompt's own
residual from before the last attention layer, and the \emph{shifted}
state, which is another prompt's.  Ordinary minus skipped isolates that
layer's final-token update; skipped minus shifted isolates where the
residual comes from.  In all four models most of the gap comes from the
residual's source, not from the attention update.

\begin{table}[htbp]
\centering
\scriptsize
\begin{tabular}{lrr}
\toprule
 & \multicolumn{2}{c}{Increase in supplied location's score relative to} \\
\cmidrule(lr){2-3}
Incoming state after the last attention layer & Memory's own answer & Recipient's answer \\
\midrule
\multicolumn{3}{l}{\textit{Nemotron-H}: both factors contribute} \\
Ordinary execution & 0.735 \(\pm\) 0.048 & 1.006 \(\pm\) 0.057 \\
Own pre-attention state, attention skipped & 0.678 \(\pm\) 0.046 & 1.241 \(\pm\) 0.075 \\
Foreign prompt's pre-attention state & 0.365 \(\pm\) 0.045 & 0.740 \(\pm\) 0.080 \\
\quad Ordinary minus own state (attention ran) & +0.058 \(\pm\) 0.014 & \(-\)0.236 \(\pm\) 0.030 \\
\quad Own minus foreign state (foreign token) & +0.313 \(\pm\) 0.050 & +0.501 \(\pm\) 0.091 \\
\quad Ordinary minus foreign state (total gap) & +0.370 \(\pm\) 0.051 & +0.266 \(\pm\) 0.081 \\
\midrule
\multicolumn{3}{l}{\textit{Nemotron-3 Nano}: the inserted token is the cause} \\
Ordinary execution & 0.177 \(\pm\) 0.026 & 0.242 \(\pm\) 0.031 \\
Own pre-attention state, attention skipped & 0.194 \(\pm\) 0.030 & 0.335 \(\pm\) 0.038 \\
Foreign prompt's pre-attention state & 0.115 \(\pm\) 0.025 & 0.244 \(\pm\) 0.036 \\
\quad Ordinary minus own state (attention ran) & \(-\)0.017 \(\pm\) 0.009 & \(-\)0.093 \(\pm\) 0.013 \\
\quad Own minus foreign state (foreign token) & +0.079 \(\pm\) 0.015 & +0.091 \(\pm\) 0.024 \\
\quad Ordinary minus foreign state (total gap) & +0.061 \(\pm\) 0.015 & \(-\)0.002 \(\pm\) 0.025 \\
\midrule
\multicolumn{3}{l}{\textit{Nemotron-Nano-9B-v2}: the inserted token is the cause} \\
Ordinary execution & 0.341 \(\pm\) 0.030 & 0.413 \(\pm\) 0.033 \\
Own pre-attention state, attention skipped & 0.397 \(\pm\) 0.030 & 0.605 \(\pm\) 0.038 \\
Foreign prompt's pre-attention state & 0.189 \(\pm\) 0.026 & 0.377 \(\pm\) 0.044 \\
\quad Ordinary minus own state (attention ran) & \(-\)0.056 \(\pm\) 0.014 & \(-\)0.192 \(\pm\) 0.019 \\
\quad Own minus foreign state (foreign token) & +0.208 \(\pm\) 0.027 & +0.229 \(\pm\) 0.046 \\
\quad Ordinary minus foreign state (total gap) & +0.152 \(\pm\) 0.027 & +0.037 \(\pm\) 0.043 \\
\midrule
\multicolumn{3}{l}{\textit{Granite-4.0-H-Small}: the inserted token is the cause} \\
Ordinary execution & 0.363 \(\pm\) 0.028 & 0.474 \(\pm\) 0.029 \\
Own pre-attention state, attention skipped & 0.409 \(\pm\) 0.032 & 0.700 \(\pm\) 0.039 \\
Foreign prompt's pre-attention state & 0.240 \(\pm\) 0.030 & 0.484 \(\pm\) 0.046 \\
\quad Ordinary minus own state (attention ran) & \(-\)0.046 \(\pm\) 0.009 & \(-\)0.226 \(\pm\) 0.017 \\
\quad Own minus foreign state (foreign token) & +0.169 \(\pm\) 0.023 & +0.215 \(\pm\) 0.040 \\
\quad Ordinary minus foreign state (total gap) & +0.123 \(\pm\) 0.024 & \(-\)0.010 \(\pm\) 0.040 \\
\bottomrule
\end{tabular}

\caption{Bypass-source experiment: transposed-minus-control change in the
supplied location's score, relative to the memory prompt's own answer and
to the recipient's answer, under three incoming states after the last
attention layer, with the paired differences between them; 95\%
identity-bootstrap intervals over 200 identities per model.}
\label{tab:bypass-source}
\end{table}

\subsection{Ordinary accuracy under recurrent-output disruption}
\label{app:dependence}

\begin{table}[htbp]
 \centering\scriptsize
 \setlength{\tabcolsep}{3pt}
 \resizebox{\linewidth}{!}{\begin{tabular}{lrrlrr}
\toprule
Condition (final token only) & Baseline (\%) & Intervened (\%) & Paired drop (points, 95\%) & Correct$\to$wrong & Wrong$\to$correct \\
\midrule
\multicolumn{6}{l}{\emph{Word wording}: source 70.7\%, recipient 62.0\%, 100 identities} \\
Post-attention outputs, calibration means (primary) & 69.8 & 67.3 & $2.5 \pm 1.9$ & 9.0 & 12.6 \\
Earlier five recurrent outputs, calibration means & 69.8 & 62.2 & $7.6 \pm 2.8$ & 17.9 & 16.2 \\
Post-attention outputs, zeros & 69.8 & 50.7 & $19.1 \pm 3.5$ & 34.1 & 15.6 \\
Earlier five recurrent outputs, zeros & 69.8 & 32.3 & $37.5 \pm 4.0$ & 61.0 & 16.9 \\
\quad Post-attention minus earlier, means & & & $-5.1 \pm 3.1$ & & \\
\midrule
\multicolumn{6}{l}{\emph{Retrospective}: memory swap, transposed minus control memory, ordinary execution, all five memories} \\
Word wording (200 identities) & 69.2 & & $-2.8 \pm 1.5$ & \multicolumn{2}{l}{Control minus unswapped $-0.2 \pm 0.5$} \\
Story wording (200 identities) & 63.3 & & $-3.6 \pm 1.4$ & \multicolumn{2}{l}{Control minus unswapped $0.0 \pm 0.3$} \\
\bottomrule
\end{tabular}
}
 \caption{Ordinary seven-candidate top-1 accuracy when final-token
 Mamba outputs are replaced, Nemotron-H, no residual transplant.  Paired
 drops are baseline minus intervened accuracy, averaged within identity
 (nine source prompts and one recipient prompt, weighted 9:1); transition
 rates are conditional on baseline correctness.  The memory-swap rows
 compare the transposed with the control memory on the same units.}
 \label{tab:dependence}
\end{table}

\paragraph{Does the ordinary answer depend on the post-attention outputs?}
A preregistered test runs the ordinary two-hop prompts with no transplant
and replaces only the final-token outputs of the five post-attention Mamba
layers with per-layer means from a separate calibration pool; the null,
identity, installed-output, and prefix-scope checks are exact.  The
replacement lowers accuracy modestly, the memory swap has an accuracy
consequence of the same order, and the same replacement at an earlier layer
costs 7.6 points against these layers' 2.5, a paired difference of
\(-\)5.1 \(\pm\) 3.1 (Table~\ref{tab:dependence}).  This is what
the ordinary forward pass predicts: the answer is already resolved before
these layers, so removing them costs little, and it is the direct evidence
for that reading, which Section~\ref{sec:concentration} reaches from the
other side by measuring that attention carries the conversion.  What Section~\ref{sec:suffix}
measures is not their necessity in an ordinary pass but what they do with
what is in their memory, which is why the read appears as a change of odds
in ordinary execution and as a change of the first-ranked answer once the
incoming state carries no resolved answer of its own
(Figure~\ref{fig:memory-read}, right).

\FloatBarrier
\section{Task variants and robustness}
\label{app:variants}

\paragraph{Task-family variants.}
Table~\ref{tab:generality} repeats the binding-task sweep under four
variants with the same seeds and candidates: entity-less filler sentences
after every fact, single-token color values in place of numbers, a one-hop
entity--attribute binding with no intermediate key, and a bAbI-style
two-fact task with typed facts and location answers.  Every hybrid run
keeps the key-dominant class at the last attention layer, the
concentration stays high under filler sentences and color values, and on
every two-hop variant the dense models stay below the hybrids.  On the
one-hop task, with no intermediate key to assemble, the rise splits over
the last two attention layers; on the word task part of it falls on the
Mamba layers after Nemotron-H's last attention layer
(Appendix~\ref{app:suffix-retrieval}).

\begin{table}[htbp]
\centering
\scriptsize
\setlength{\tabcolsep}{3pt}
\resizebox{\textwidth}{!}{%
\begin{tabular}{llrrllll}
\toprule
Variant & Model & Gate (\%) & Site & \(\kappa\) & \(\pi\) & Class & \(C\) \\
\midrule
Filler context & Nemotron-H-8B & 65/66 & pre-L40 & 0.81 \(\pm\) 0.12 & 0.36 \(\pm\) 0.09 & Key-dominant (dual) & 0.90 \(\pm\) 0.02 \\
Filler context & Qwen2.5-7B & 66/66 & pre-L22 & 0.53 \(\pm\) 0.13 & 0.53 \(\pm\) 0.13 & Balanced & 0.52 \(\pm\) 0.02 \\
Filler context & Llama-3.1-8B & 83/86 & pre-L16 & 0.78 \(\pm\) 0.10 & 0.37 \(\pm\) 0.08 & Key-dominant (dual) & 0.35 \(\pm\) 0.01 \\
Word-valued attributes & Nemotron-H-8B & 70/77 & pre-L40 & 0.82 \(\pm\) 0.12 & 0.20 \(\pm\) 0.08 & Key-dominant (dual) & 0.89 \(\pm\) 0.01 \\
Word-valued attributes & Granite-H & 66/62 & pre-L35 & 0.66 \(\pm\) 0.09 & 0.10 \(\pm\) 0.06 & Key-dominant (dual) & 0.89 \(\pm\) 0.04 \\
Word-valued attributes & Qwen2.5-7B & 66/68 & pre-L21 & 0.39 \(\pm\) 0.10 & 0.33 \(\pm\) 0.08 & Balanced & 0.32 \(\pm\) 0.01 \\
Word-valued attributes & Llama-3.1-8B & 85/80 & pre-L16 & 0.68 \(\pm\) 0.07 & 0.32 \(\pm\) 0.06 & Key-dominant (dual) & 0.26 \(\pm\) 0.01 \\
One-hop binding & Nemotron-H-8B & 100/100 & pre-L40 & 3.17 \(\pm\) 0.24 & 1.39 \(\pm\) 0.21 & Key-dominant (dual) & 0.50 \(\pm\) 0.01 \\
One-hop binding & Granite-H & 100/100 & pre-L35 & 2.85 \(\pm\) 0.18 & 0.41 \(\pm\) 0.10 & Key-dominant (dual) & 0.57 \(\pm\) 0.01 \\
One-hop binding & Qwen2.5-7B & 100/100 & pre-L22 & 3.70 \(\pm\) 0.39 & 3.75 \(\pm\) 0.39 & Dual, unbalanced & 0.61 \(\pm\) 0.01 \\
One-hop binding & Llama-3.1-8B & 100/100 & pre-L16 & 3.39 \(\pm\) 0.26 & 1.87 \(\pm\) 0.22 & Key-dominant (dual) & 0.33 \(\pm\) 0.01 \\
bAbI-style two-fact & Nemotron-H-8B & 70/72 & pre-L40 & 1.75 \(\pm\) 0.19 & 0.58 \(\pm\) 0.14 & Key-dominant (dual) & 0.66 \(\pm\) 0.02 \\
bAbI-style two-fact & Qwen2.5-7B & 77/77 & pre-L22 & 1.80 \(\pm\) 0.23 & 1.36 \(\pm\) 0.21 & Key-dominant (dual) & 0.32 \(\pm\) 0.01 \\
bAbI-style two-fact & Llama-3.1-8B & 70/62 & pre-L17 & 1.19 \(\pm\) 0.14 & 0.41 \(\pm\) 0.12 & Key-dominant (dual) & 0.37 \(\pm\) 0.02 \\
bAbI-style, numeric values & Nemotron-H-8B & 90/86 & pre-L40 & 2.60 \(\pm\) 0.19 & 0.78 \(\pm\) 0.16 & Key-dominant (dual) & 0.83 \(\pm\) 0.01 \\
\bottomrule
\end{tabular}
}
\caption{Task-family variants of the binding task.  Gate is source/recipient
seven-way top-1; the site is the last attention layer for the sequential
hybrids and the key peak otherwise; \(\kappa\) and \(\pi\) carry 97.5\%
intervals and \(C\) a 95\% interval.  Runs below the accuracy gate are absent.}
\label{tab:generality}
\end{table}

\FloatBarrier
\section{Statistical procedures, thresholds, and multiplicity}
\label{app:stats}

\paragraph{Preregistration and statistics.}
For every confirmatory experiment the construction, site, sample, and
numerical thresholds were written to a configuration file before the
held-out sample was run; we call such thresholds \emph{prespecified}.  We
write \(e\pm h\) for an estimate \(e\) whose interval has half-width
\(h\); every decision uses the interval's own bounds, which are close to
symmetric.  Intervals are 95\% percentile bootstrap intervals over 10,000
replicates clustered by identity, except that the factorial classification
uses 97.5\% intervals for its two contrasts and a 90\% interval for the
equivalence test.  Head-set and path comparisons use fixed random banks with
add-one randomization or Holm correction.  Every run passes exact checks:
transplanting a recipient's own state or a zero patch changes no logit, and
\(R_{\rm by}+P_{\rm all}=F\) holds exactly.

\subsection{Threshold sensitivity and multiplicity}
\label{app:sensitivity}

A site is \emph{dual-supported} when the 97.5\% lower bounds of both
\(\kappa\) and \(\pi\) exceed zero; \emph{key-dominant} when the 97.5\%
lower bounds of \(\kappa\) and of \(\kappa-\pi\) exceed zero (with dual
support when \(\pi\)'s does as well), and position-dominant in the mirror
case.  It is \emph{balanced} when it is dual-supported and the 90\%
interval of \(\kappa-\pi\) lies inside \(\pm0.30\) logits.  Every threshold
was fixed before the confirmatory factorial runs.

The rule of Section~\ref{sec:constructions} has three fixed constants: the
lower-bound level for \(\kappa\), \(\pi\), and \(\kappa-\pi\), the interval
level for the balance test, and the balanced band.
Table~\ref{tab:sensitivity} re-classifies every reference site under 18
combinations of these constants and counts the layers of each profile whose
class changes.  Every key-dominant site keeps its class under every
setting; the only changes are two balanced sites at the edge of the band
(Qwen2.5-7B pre-L22 and Falcon-H1-7B pre-L29), and few layers of any
profile change class.

\begin{table}[htbp]
\centering
\scriptsize
\setlength{\tabcolsep}{4pt}
\resizebox{\textwidth}{!}{%
\begin{tabular}{llrlll}
\toprule
Model & Task & Site & Preregistered class & Classes across the grid & Layers changed (max / mean) \\
\midrule
Nemotron-H-8B & Arrow & pre-L40 & Key-dominant (dual) & Key-dominant (dual) & 4 / 1.3 of 52 \\
Nemotron-H-8B (pre-L39) & Arrow & pre-L39 & Key-dominant (dual) & Key-dominant (dual) & 4 / 1.3 of 52 \\
Nemotron-H-8B & Colon & pre-L40 & Key-dominant (dual) & Key-dominant (dual) & 0 / 0.0 of 2 \\
Nemotron-H-8B & Binding & pre-L40 & Key-dominant (dual) & Key-dominant (dual) & 0 / 0.0 of 52 \\
Nemotron-H-4B & Arrow & pre-L40 & Key-dominant (dual) & Key-dominant (dual) & 0 / 0.0 of 52 \\
Nemotron-H-4B & Binding & pre-L40 & Key-dominant (dual) & Key-dominant (dual) & 4 / 1.3 of 52 \\
Granite-H & Arrow & pre-L35 & Key-dominant (dual) & Key-dominant (dual) & 3 / 1.7 of 40 \\
Granite-H & Binding & pre-L35 & Key-dominant (dual) & Key-dominant (dual) & 1 / 0.7 of 40 \\
Qwen2.5-7B & Arrow & pre-L22 & Balanced & Balanced, dual, unbalanced & 4 / 1.7 of 28 \\
Qwen2.5-7B (pre-L26) & Arrow & pre-L26 & Unresolved & Unresolved & 5 / 2.2 of 28 \\
Qwen2.5-7B & Binding & pre-L22 & Balanced & Balanced & 1 / 0.7 of 28 \\
Qwen2.5-14B & Arrow & pre-L36 & Position-dominant (dual) & Position-dominant (dual) & 3 / 1.0 of 48 \\
Qwen2.5-14B & Binding & pre-L36 & Balanced & Balanced & 2 / 1.0 of 48 \\
Mistral-7B & Arrow & pre-L19 & Key-dominant (dual) & Key-dominant (dual) & 2 / 1.3 of 32 \\
Mistral-7B & Binding & pre-L19 & Key-dominant (dual) & Key-dominant (dual) & 0 / 0.0 of 32 \\
Llama-3.1-8B & Arrow & pre-L16 & Key-dominant (dual) & Key-dominant (dual) & 2 / 0.7 of 32 \\
Llama-3.1-8B & Binding & pre-L16 & Key-dominant (dual) & Key-dominant (dual) & 2 / 0.7 of 32 \\
Gemma-2-9B & Arrow & pre-L28 & Key-dominant (dual) & Key-dominant (dual) & 1 / 0.3 of 42 \\
Gemma-2-9B & Binding & pre-L28 & Key-dominant (dual) & Key-dominant (dual) & 2 / 1.0 of 42 \\
Falcon-H1-3B & Arrow & pre-L21 & Key-dominant (dual) & Key-dominant (dual) & 2 / 0.7 of 32 \\
Falcon-H1-3B & Binding & pre-L21 & Key-dominant (dual) & Key-dominant (dual) & 4 / 2.3 of 32 \\
Falcon-H1-7B & Arrow & pre-L29 & Balanced & Balanced, key-dominant (dual), dual, unbalanced & 2 / 1.2 of 44 \\
Falcon-H1-7B & Binding & pre-L29 & Balanced & Balanced & 3 / 1.7 of 44 \\
\bottomrule
\end{tabular}
}
\caption{Threshold sensitivity of the classification.  ``Site'' is the
last attention layer for sequential hybrids and the key peak
otherwise; ``classes across the grid'' lists every class the site receives
over the 18 settings; ``layers changed'' is the maximum and mean number of
layers in the depth profile whose class differs from the preregistered
setting.  Binding-task rows are absent for models that failed the binding
gate.}
\label{tab:sensitivity}
\end{table}

\paragraph{The concentration against the number of attention layers.}
\(C=\max_l\Delta_l^+/\sum_l\Delta_l^+\), with \(\Delta_l^+=\max(\Delta_l,0)\),
divides the largest positive increment by the sum over all layers, so a
model with few attention layers has fewer places for the conversion than a
dense model with attention in every layer.
Table~\ref{tab:concentration-null} therefore restricts the statistic to
attention-containing layers (\(C_{\rm attn}\)), compares it with the uniform
null over those layers, and gives the effective attention count \(e^{H}\).
Against that null the ordering reverses: the dense models are the
concentrated ones, because they spread the rise over a small part of the
attention they have while the hybrids use nearly all of theirs.  The raw
comparison depends strongly on how many attention layers are available;
these are observational comparisons of different pretrained models, not a
matched test of changing the architecture.

\begin{table}[htbp]
\centering
\scriptsize
\setlength{\tabcolsep}{4pt}
\begin{tabular}{lrrrrrrr}
\toprule
 & & \multicolumn{3}{c}{Arrow task} & \multicolumn{3}{c}{Binding task} \\
\cmidrule(lr){3-5}\cmidrule(lr){6-8}
Model & Attn.\ blocks & \(C_{\rm attn}\) & \(C_{\rm attn}/(1/n)\) & \(e^{H}\) & \(C_{\rm attn}\) & \(C_{\rm attn}/(1/n)\) & \(e^{H}\) \\
\midrule
Nemotron-H-8B & 4 & 0.71 & 2.9 & 1.9 & 0.72 & 2.9 & 1.9 \\
Nemotron-H-4B & 4 & 0.66 & 2.7 & 1.9 & 0.71 & 2.8 & 1.8 \\
Granite-H & 4 & 0.73 & 2.9 & 1.8 & 0.57 & 2.3 & 2.0 \\
RecurrentGemma-9B & 12 & 0.80 & 9.6 & 2.0 & --- & --- & --- \\
Nemotron-3-Nano-30B & 6 & 0.78 & 4.7 & 1.7 & 0.75 & 4.5 & 1.9 \\
\midrule
Falcon-H1-3B & 31 & 0.59 & 18.3 & 4.6 & 0.56 & 17.4 & 4.8 \\
Falcon-H1-7B & 43 & 0.70 & 29.9 & 3.2 & 0.73 & 31.3 & 3.0 \\
\midrule
Qwen2.5-7B & 27 & 0.61 & 16.5 & 2.9 & 0.54 & 14.5 & 3.2 \\
Qwen2.5-14B & 47 & 0.38 & 18.0 & 5.5 & 0.36 & 16.8 & 5.9 \\
Mistral-7B & 31 & 0.44 & 13.8 & 3.4 & 0.43 & 13.4 & 4.0 \\
Llama-3.1-8B & 31 & 0.35 & 11.0 & 7.1 & 0.32 & 10.0 & 8.0 \\
Gemma-2-9B & 41 & 0.65 & 26.6 & 4.1 & 0.68 & 27.9 & 3.6 \\
\bottomrule
\end{tabular}

\caption{Resolution concentration restricted to attention-containing
layers on the donor-answer effect.  \(n\) counts attention-containing
layers with a measured increment (the final layer has none);
\(C_{\rm attn}/(1/n)\) is the ratio to the uniform null over those layers;
\(e^{H}\) is the effective attention count carrying the
positive increments.}
\label{tab:concentration-null}
\end{table}

\paragraph{Multiplicity across layers.}
The depth profiles are descriptive.  Under a Bonferroni correction of both
interval levels across every swept layer, the sequential hybrids keep their
key-dominant band through the last attention layer and lose at most an
edge layer, while single-layer bands in dense and parallel models do not
survive the correction.

\FloatBarrier
\section{Alternative explanations, boundary conditions, and controls}
\label{app:alternatives}

\paragraph{Intervention checks.}
The shifted memory experiments show no excess final-layer tail atypicality
relative to unmodified recipients in either wording, and every memory-swap
run passes its decoder and accuracy checks (Appendix~\ref{app:memory}).
These diagnostics characterize the interventions without certifying that
every inserted state is on-distribution.

Related studies locate retrieval and answer-selection mechanisms in dense
and recurrent models
\citep{lieberum2023doescircuitanalysisinterpretability,wiegreffe2025answer,sharma2024locating,ensign2024investigating,wang-etal-2026-understanding,zani2025contextual,wu2025retrieval}.
Our hybrid panel uses architectures that combine state space computation
with attention
\citep{gu2024mamba,dao2024transformersssmsgeneralizedmodels,waleffe2024empiricalstudymambabasedlanguage,nvidia2025nemotronhfamilyaccurateefficient}.
We also run path
patching \citep{goldowskydill2023localizingmodelbehaviorpath} and causal
scrubbing \citep{chan2022causal}
and, because interventions can create atypical states
\citep{makelov2024is,grant2026addressing}, track the distance of intervened
trajectories from natural activations.

\begin{table}[htbp]
\centering
\scriptsize
\setlength{\tabcolsep}{3pt}
\begin{tabular}{
  >{\raggedright\arraybackslash}p{0.30\linewidth}
  >{\raggedright\arraybackslash}p{0.64\linewidth}}
\toprule
Alternative & Evidence against it \\
\midrule
Fixed donor answer &
One donor tracks three recipient bindings on 81\% of crosses; \(S_A\) scrub destroys the effect \\
Recipient keeps its own answer \(A_r\) &
\(\Delta\ell(B_j)-\Delta\ell(A_r)=4.17\) logits \\
Query token, not key, is transferred &
Recipient maps \(q_j\) away from \(B_j\) in 97\% of cells; excluding the rest leaves \(\kappa\) unchanged \\
State is only a row pointer &
The key beats the position by 0.77 logits with a reference baseline; \(S_P\) scrub destroys the effect \\
Any nearby site works &
Pre-L39 whole residual carries three quarters of the effect, same class \\
Any eight heads carry it &
Top-8 path exceeds all 255 random eight-head paths \\
Heads merely relay donor outputs &
Donor-clamped heads give negative key selectivity; \(F-D_{\rm attn}>0\) \\
The site is special only because we nominated it &
Full depth sweeps of twelve models with no outcome selection; the three frozen sites keep their class \\
The result depends on the prompt format &
Second surface form keeps the class at pre-L39 and pre-L40 \\
The state barely encodes the key &
Ridge probe retrieves the key on 78\% of donors at pre-L40 (3-way) \\
Suffix clamp only causes nonspecific damage &
Nominated-pair drop exceeds comparison-pair drop; same-key cross-donor rescue exceeds different-key control (Table~\ref{tab:suffix-rescue}) \\
Post-attention layers only pass along an answer &
Memory prompts with the same own answer but transposed bindings redirect the lookup in four hybrids, in ordinary execution as well; the carried-answer control is zero in Nemotron-H and Nano (Table~\ref{tab:memory-map}) \\
Nano's later lookup has no identified supplier &
L37 interruption, specificity, rescue, and increased late-head dependence (Table~\ref{tab:late-supply}) \\
\bottomrule
\end{tabular}
\caption{Alternative accounts and the evidence against each.}
\label{tab:alternatives}
\end{table}

\FloatBarrier

\end{document}